\documentclass[10pt,twocolumn,letterpaper]{article}

\usepackage{cvpr}
\usepackage{times}
\usepackage{epsfig}
\usepackage{graphicx}
\usepackage{amsmath}
\usepackage{amssymb}
\usepackage{bm}
\usepackage{multirow}
\usepackage{cite}
\usepackage{gensymb}
\usepackage{rotating}
\usepackage{array}

\newcolumntype{?}{!{\vrule width 2pt}}

\newcommand{\ts}[1]{\textcolor{blue}{Torsten: #1}\xspace} 

\newcommand{\ms}[1]{\textcolor{red}{M\r{a}ns: #1}\xspace} 

\usepackage[pagebackref=true,breaklinks=true,letterpaper=true,colorlinks,bookmarks=false]{hyperref}

\cvprfinalcopy 

\def\cvprPaperID{5332} 

\newcommand{\PAR}[1]{\vskip4pt \noindent{\bf #1~}}

\ifcvprfinal\fi
\begin{document}

\title{A Cross-Season Correspondence Dataset for Robust Semantic Segmentation}

\author{M\r{a}ns Larsson$^{1}$ \hspace{2mm} Erik Stenborg$^{1,3}$ \hspace{2mm} Lars Hammarstrand$^1$ \hspace{2mm} Torsten Sattler$^1$ \hspace{2mm} Marc Pollefeys$^{2,4}$ \\ Fredrik Kahl$^1$ \\
$^1$Chalmers University of Technology \hspace{5mm}  $^2$Department of Computer Science, ETH Zurich\\
$^3$ Zenuity \hspace{5mm} $^4$ Microsoft \\ 
}

\maketitle

\begin{abstract}
In this paper, we present a method to utilize 2D-2D point matches between images taken during different image conditions to train a convolutional neural network for semantic segmentation. Enforcing label consistency across the matches makes the final segmentation algorithm robust to seasonal changes. We describe how these 2D-2D matches can be generated with little human interaction by geometrically matching points from 3D models built from images. Two cross-season correspondence datasets are created providing 2D-2D matches across seasonal changes as well as from day to night. The datasets are made publicly available to facilitate further research. We show that adding the correspondences as extra supervision during training improves the segmentation performance of the convolutional neural network, making it more robust to seasonal changes and weather conditions. 
\end{abstract}

\section{Introduction}
Semantic segmentation is the task of assigning a class label to each pixel in an image and is one of the fundamental problems in computer vision. 
Semantic segmentation has also been used to integrate  higher-level scene understanding into other computer vision problems, \eg,  
dense 3D reconstruction~\cite{Cherabier2018ECCV,Blaha2016CVPR,Cherabier20163DV,Haene2017PAMI,Haene2014CVPR,Kundu2014ECCV,Savinov2016CVPR,Schneider2016IV,Song2017CVPR}, SLAM~\cite{Lianos2018ECCV,Bowman2017ICRA} and Structure-from-Motion~\cite{Bao2011CVPR}, 
%
3D model alignment~\cite{Yu2015CVPR,Cohen2016ECCV,Cohen2015ICCV}, and  location recognition~\cite{Arandjelovic2014ACCV,MousavianICRA2015,Singh2016LSVGL}. 
%

Visual localization is the problem of estimating the camera pose of an image~\cite{Brachmann2018CVPR,Li2012ECCV}, typically from a set of 2D-3D matches between image pixels and 3D scene points. 
In the context of long-term visual localization~\cite{Toft2017ICCVW,Toft2018ECCV,Schoenberger2018CVPR,Yu2018IROS,Stenborg2018LongTermVL}, semantics are proven useful to be able to handle variations in scene geometry and appearance, \eg, due to seasonal changes. 
These methods are based on the idea that the semantic meaning of a scene part is invariant to such changes. 
Semantics are thus used to establish the 2D-3D matches required for pose estimation when matching solely based on image appearance fails.

As of yet, however, the assumption that the same semantic segmentation can be reliably reproduced  under different conditions does not hold. 
This is mainly due to that the labeled datasets used to train the semantic segmentation algorithm itself are limited to only a few conditions. 
As the pixel-level annotations are performed by hand, adding more training data is both time consuming and expensive~\cite{Cordts2016Cityscapes,neuhold2017mapillary}.
However, even rather noisy segmentations improve localization performance compared to not using semantic information~\cite{Schoenberger2018CVPR,Toft2018ECCV}. 
This naturally leads to the question whether it is possible to build a feedback loop: 
Can semantic segmentation algorithms be improved via visual localization? 
Can this in turn lead to more robust localization results? 
\begin{figure}
    \centering
    \includegraphics[width=0.47\textwidth]{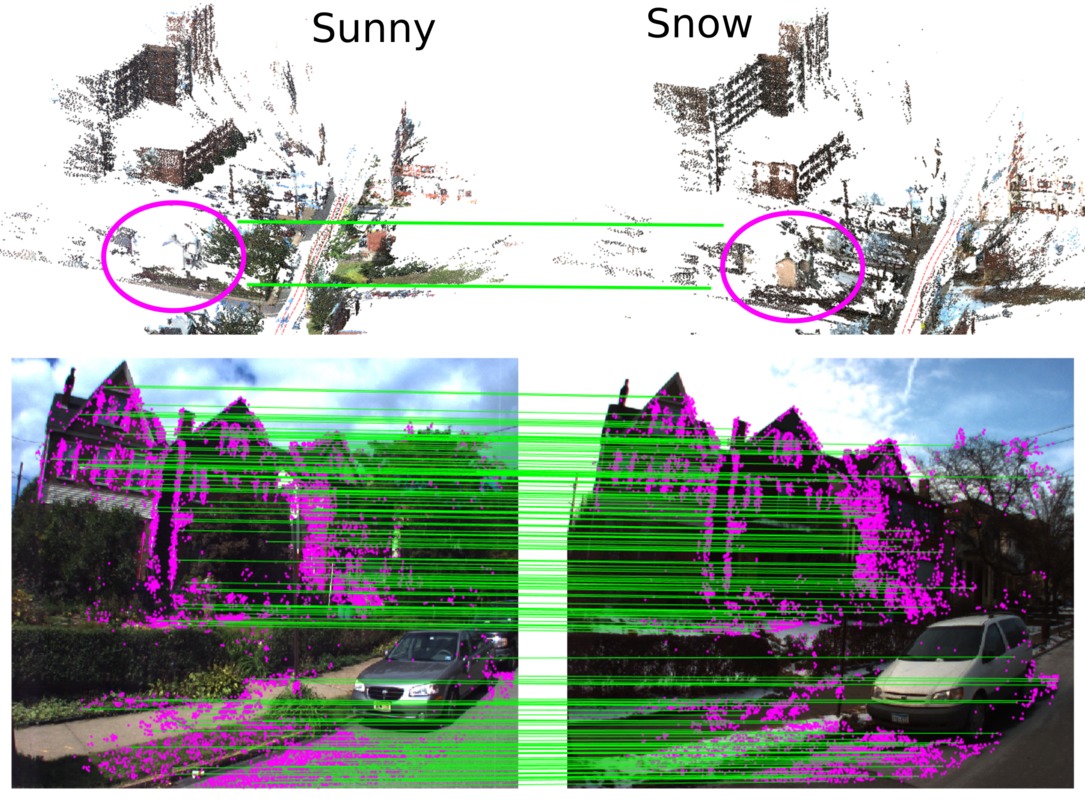}
    \caption{2D-2D matches between images taken under different conditions, established using matches against a 3D point cloud.}
    \label{fig:intr}
\end{figure}

As part of estimating the camera pose of an image taken under one condition against a 3D model built from images, a set of 2D-3D matches is established. 
Usually, the reference 3D model is built from images~\cite{sattler2018benchmarking}. 
Thus, the 2D-3D correspondences lead to a set of 2D-2D matches between images taken under different conditions, as depicted in Fig.~\ref{fig:intr}. 
In this paper, we show that these pixel-level matches can be used to improve semantic segmentation algorithms. 

In detail, this paper makes the following contributions: 
\textbf{1}) A 2D-2D  match between two images taken under different conditions provides a constraint on the training process, namely that these two pixels should receive the same label. 
We use this insight to formulate a loss function that can be added to the training process, without the need to modify the  architecture of the segmentation algorithm. 
\textbf{2}) We show that the required set of correspondences can be generated with little human supervision, albeit without ground truth labels. 
This is in stark contrast to creating labeled training datasets, where significant human effort is required~\cite{richter2016playing}. 
We make our cross-season correspondence dataset publicly available\footnote{\url{https://visuallocalization.net}}. 
\textbf{3}) We show that using our correspondence-based loss, together with a few coarsely annotated images required to prevent trivial solutions, can lead to significant improvements in segmentation quality in the context of changing imaging conditions. 
%
The improvements are especially significant if the base training set covers only a single condition. 
%
%




\section{Related Work}

\PAR{Semantic Segmentation.} 
The performance of semantic segmentation algorithms has seen a large increase during the last few years based on the advances of deep neural networks. The seminal work of Long \etal \cite{long2015fully} showed that convolutional neural networks (CNNs), initially trained for classification, can be transformed to fully convolutional networks (FCNs) for semantic segmentation. Follow up work has improved upon FCNs by for example enlarging the receptive field \cite{chen2018deeplab, yu2015multi}, incorporating higher level context \cite{zhao2017pyramid} or fusing multi-scale features \cite{chen2016attention, ronneberger2015u}. In addition, the combination of FCNs and structure models, such as conditional random field (CRF), have been thoroughly studied, either as a post-processing step \cite{chen2018deeplab} or as part of the network \cite{zheng2015conditional, liu2015semantic, larsson2018revisiting}, enabling end-to-end training.

Training these networks requires a large amount of annotated images, which for semantic segmentation can be costly and time-consuming to acquire. In response many weakly supervised approaches have been suggested utilizing labels in the form of bounding boxes \cite{dai2015boxsup, khoreva2017simple, papandreou2015weakly}, image level tags \cite{papandreou2015weakly, pathak2015constrained, Pinheiro_2015_CVPR, souly2017semi} or points \cite{bearman2016s}. Similarly, we improve the segmentation performance of a FCN by utilizing supervision that requires less manual effort to acquire than pixel-level annotations. However, instead of using weaker, but still manually annotated, labels we propose the use of data that can be acquired in a semi-automatic fashion.

\PAR{Domain Adaptation.}
Domain adaptation methods aim at learning a model that performs well in the target domain, given that there are only available annotations in the source domain.
Early work 
includes~\cite{kulis2011you, saenko20adapting} that transforms the features to either a domain invariant feature space~\cite{saenko20adapting} or the source feature domain~\cite{kulis2011you}. Several works have focused on domain adaptation for CNN models \cite{ganin2016domain, long2015learning, long2016unsupervised, tzeng2017adversarial}. These methods learn models that produce domain invariant features, either by aligning the target and feature distributions \cite{long2015learning,long2016unsupervised} or by using an adversarial training setup, encouraging domain confusion \cite{ganin2016domain, tzeng2017adversarial}.

Recently, several domain adaptation methods for dense prediction tasks have been presented~\cite{hoffman2016fcns, tsai2018learning, wulfmeier2017addressing, wulfmeier2018incremental, chen2018road, zou2018unsupervised, sankaranarayanan2018learning}. Most of these \cite{hoffman2016fcns, murez2017image, hoffman2017cycada, tsai2018learning, chen2018road, zou2018unsupervised, sankaranarayanan2018learning} use synthetic datasets, \eg, \cite{ros2016synthia,richter2016playing}, enabling automatic generation of large amount of annotated synthetic images. The methods presented in \cite{sankaranarayanan2018learning, murez2017image, hoffman2017cycada} all use some form of image translation method such as to transform the source images into the target domain before performing segmentation. Another common approach is to use an adversarial training setting as in \cite{wulfmeier2017addressing, wulfmeier2018incremental, hoffman2016fcns} where the network is encouraged to produce features that fool a domain discriminator. 

We are interested in increasing the performance of our segmentor on images different from the source domain. Specifically, the different conditions and seasons included in our correspondences can be seen as target domains. Rather than using unsupervised domain adaption, we use 3D geometric consistency as a supervisory signal. Our cross-season correspondence datasets facilitate the adaption of the segmentation method across the different target domains, removing the need to rely on purely unsupervised domain adaptation methods.

\PAR{Semantic 3D Mapping.}
Semantic 3D reconstruction approaches~\cite{mccormac2017semanticfusion,Cherabier2018ECCV,Blaha2016CVPR,Cherabier20163DV,Haene2017PAMI,Haene2014CVPR,Kundu2014ECCV,Savinov2016CVPR,Schneider2016IV,Song2017CVPR} uses semantic image segmentations to aid the reconstruction process. 
They use a voxel volume to represent the scene and jointly reason about geometric and semantic occupancy. 
Using semantics typically leads to more consistent and complete 3D models. 
These 3D models can be projected into the images to obtain refined semantic segmentations~\cite{Haene2017PAMI}. 
However, the semantic reconstruction process is significantly more complex compared to the multi-view stereo process we use to obtain correspondences. 
Methods exists that jointly predict depth and semantics~\cite{Ladicky2012IJCV,Eigen2015ICCV} or use depth information to aid semantic segmentation~\cite{hazirbas2016fusenet}. 
Yet, they still rely on labelled data. 
We are not aware of any work using pixel correspondences obtained via 3D models to create additional constraints for semantic segmentation.


\PAR{Datasets for 3D and semantics.} Large databases for indoor~\cite{Xiao2013Sun3D,Chang20173DV,Dai2017CVPR} and outdoor scenes~\cite{Geiger2013Kitti,apolloscape_arXiv_2018} provide both semantics and 3D geometry and can thus also be used to semantically annotated images via geometry~\cite{Xie2016CVPR}. Yet, we are not aware of any such dataset that captures different seasonal and illumination conditions. In this sense, our work closes a gap in the literature.
\section{Semantic Correspondence Loss \label{sec:loss}}
The rationale behind using 2D-2D image correspondences is that the CNN initially, being pre-trained on a large-scale dataset such as Cityscapes~\cite{Cordts2016Cityscapes}, performs well on images taken during favourable conditions (\ie similar conditions as in the training set). We can then use the correspondence data to enforce labeling consistency between images captured during favourable conditions and images captured during challenging conditions. To this end we define and test two different loss functions based on the hinge loss and the cross-entropy loss that will encourage labeling consistency. The losses are designed for a CNN where the value of intermediate feature layers can be extracted and where the final output is an estimate of the probability distribution over the class labels for each input pixel. 

We denote the content of one sample from the cross-season correspondence dataset as $(I^{r}, I^{t}, \bm{x}^r, \bm{x}^t)$. Here $I^{r}$ is an image from the reference traversal, $I^{t}$ an image from the target traversal, and $\bm{x}^r$ as well as $\bm{x}^t$ are the pixel positions of the matched points in the reference and target images, respectively. The reference traversal is chosen as the one with images captured during the most favourable image condition. Note that the reference images are taken from the same traversal while the target images vary between all other available traversals. 
The correspondence loss function $\mathcal{L}_{corr}$ will be a sum over all such samples
\begin{equation}
\mathcal{L}_{corr}
= \sum_{(r,t)} l(I^{r}, I^{t}, \bm{x}^r, \bm{x}^t) \enspace,
        \label{eq:lossloss}
\end{equation}
where $l$ is a hinge loss $l_{hinge}$ or a cross-entropy loss $l_{CE}$ as presented below.

Let $\bm{d}_x \in \mathbb{R}^F$ denote a feature vector of the segmentation CNN of length $F$ at pixel position $x$. This can be either the last layer of the network, where $F$ equals the number of classes or an earlier, intermediate feature layer. We define the correspondence hinge loss $l_{hinge}$ for one sample as
\begin{equation}
    l_{hinge} = \frac{1}{N}\sum_{i=1}^N \max \left ( 0, m - \frac{\bm{d}^T_{x^r_i} \bm{d}_{x^t_i}}{\| \bm{d}_{x^r_i}\| \|\bm{d}_{x^t_i}\|} \right) \enspace ,
    \label{eq:hinge}
\end{equation}
where $m$ is a margin parameter and $N$ is the number of corresponding points. 
The loss will encourage the feature vectors $\bm{d}_{x^r_i}$ and $\bm{d}_{x^t_i}$ to align up to a certain angle depending on $m$. In the experiments, we have found empirically that setting $m=0.8$ (approximately $37\degree$) works well.


For the correspondence cross-entropy loss $l_{CE}$, we begin by taking the argument of the maximum of the final feature map, \ie the most likely class, of the reference image. 
By describing the most likely class for a pixel at position $x_i$ using an one-hot encoding vector $\bm{c}_{x_i}$, the loss can be written as
\begin{equation}
    l_{CE} = -\frac{1}{N}\sum_{i=1}^N \bm{c}^T_{x^r_i} \log \left ( \bm{d}_{x^t_i}\right) \enspace,  
    \label{eq:CE}
\end{equation}
where $\log(\cdot)$ is taken element-wise. The loss will encourage the pixels in the target image to have the same labels as the corresponding pixels in the reference image. 

During training, we minimize a loss consisting of one term for the fully supervised data based on standard cross-entropy $\mathcal{L}_{sup}$ as well as one correspondence term $\mathcal{L}_{corr}$. 
The resulting overall loss is $\mathcal{L} = \mathcal{L}_{sup} + \lambda \mathcal{L}_{corr}$, where $\lambda$ is a weighting term for the impact of the correspondences. 

\section{A Cross-Season Correspondence Dataset}
This section describes the creation and the content of the cross-season correspondence dataset. Each sample of the dataset contains two nearby images taken during different seasons or weather conditions as well as a set of 2D-2D point correspondences between the images. The correspondences are automatically established using geometric 3D consistency between the two points. Geometry is typically more stable than for instance photometric information across the different conditions.

\begin{figure*}
    \centering
    \includegraphics[height=0.20\linewidth]{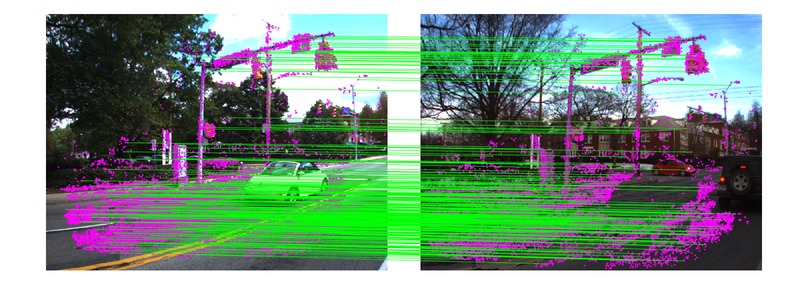}
    \includegraphics[height=0.20\linewidth]{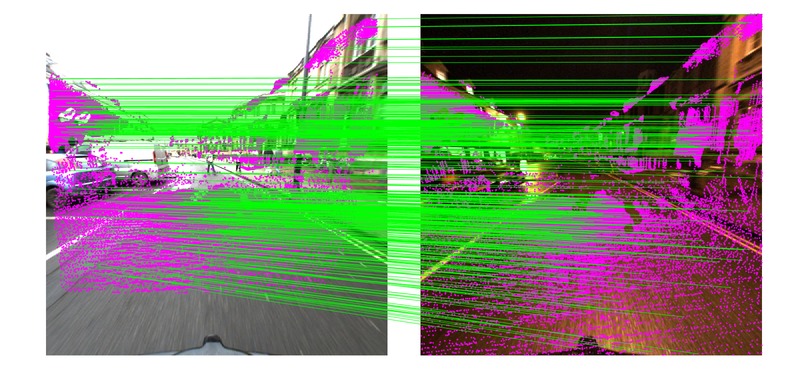}
    \caption{Visualization of a sample from our cross-season correspondence datasets. Left: CMU, right: RobotCar. Each purple point marks the pixel position of a correspondence in the  respective images. Green lines are drawn to the matching point in the other image. Note that lines are only drawn for every 50 point pair to avoid cluttering the images.}
    \label{fig:corr-ex}
\end{figure*}

A visualization of a few samples can be seen in Fig.~\ref{fig:corr-ex}. Using the datasets presented in~\cite{sattler2018benchmarking} as a starting point, we create two correspondence datasets. The datasets from~\cite{sattler2018benchmarking} used were originally based on the CMU Visual Localization dataset \cite{badino2011visual} and the the RobotCar dataset \cite{maddern20171} respectively.

The creation of the correspondence dataset can be divided into four main steps. Firstly, the camera poses for all images in all conditions need to be calculated in a common coordinate system. In our case these were kindly provided by the authors of \cite{sattler2018benchmarking}. Secondly, a dense 3D point cloud of the surrounding geometry is created, individually for each condition and traversal. Thirdly, the 3D point clouds are matched across the conditions. Since the point clouds share the same coordinate system, this matching can be done using the position of the 3D points. This removes the need for feature descriptors which might change substantially across the different conditions. Lastly, given the 3D point cloud matches, the pixel positions for the 2D-2D correspondences in each image can be calculated using the known camera positions. Each step will be detailed separately for the two datasets below.

\subsection{CMU Seasons Correspondence Dataset}
The CMU Visual Localization dataset \cite{badino2011visual} was collected over a period of 12 months in Pittsburgh, USA. A vehicle with two cameras facing forward/left and forward/right drove a 8.5 $km$ long route through central and suburban areas. We use the camera poses from twelve traversals during different seasons and weather condition available in the CMU Seasons dataset \cite{sattler2018benchmarking}. The camera poses have been calculated using bundle adjustment of SIFT points and some manually annotated image correspondences across different traversals, for further details we refer to \cite{sattler2018benchmarking}. This method gives accurate camera positions, where the position error is estimated to be under 0.10 $m$.

To create the dense point cloud for each traversal, we use the Multi-View Stereo (MVS) pipeline presented in \cite{schoenberger2016mvs}.
The MVS pipeline pipeline works in two steps. Firstly, depth and normal information is estimated using geometric and photometric information. Secondly, the depth and normal maps are fused forming a dense point cloud. This was done using the software Colmap \cite{schoenberger2016sfm, schoenberger2016mvs} with default settings.
Examples of 3D point clouds from the CMU Seasons dataset can be seen in Fig.~\ref{fig:intr}.


To find correspondences between images from two different traversals, we perform geometric matching between the two corresponding 3D point clouds. Given a reference and a target traversal, we take the two corresponding dense 3D point clouds and match points using the Euclidean distance: For each point in the first point cloud, we search for the nearest neighboring point in the other point cloud and vice versa. We establish a correspondence between two points if they are mutual nearest neighbors. 

For the resulting matches, we check from which cameras each matched point was triangulated during MVS, giving us the cameras where the point is visible.
We then go through each camera pair, where one is from the reference  and one from the target traversal and investigate the number of common 3D points that can be seen in the cameras. For pairs with at least 500 common points and where the distance between the two cameras is less than $0.5$ $m$, we do an additional pruning step to get rid of poor matches. The pruning is necessary since the first 3D matching step does not enforce any constraint that the points need to be close too each other. In addition, enforcing a closeness constraint for each camera pair enables us to use a distance threshold that depends on the distance between the point and the cameras. This is beneficial since 3D points close to the camera usually are reconstructed with greater precision. For a match to be kept, the distance between the two points must be below a threshold that depends on the distance between the point and the camera as follows 
\begin{equation}
    ||X_1 - X_2|| < \kappa D \enspace .
\end{equation}
Here $X_1$ and $X_2$ are the positions of the matched 3D points in the reference and target traversals, respectively, $D$ is the distance from the camera center to $X_1$ and $\kappa$ is a unitless parameter set to $0.01$. This means that points that are 10 $m$ from the camera need to be closer than 0.1 $m$ to be kept as a correspondence.
Table~\ref{tab:cmustat} provides a summary of the content of the final CMU Cross-Season Correspondence dataset.

\begin{table}
    \setlength{\tabcolsep}{4pt}
    \centering
    \footnotesize{
    \begin{tabular}{c|c|c}
    Condition & Image pairs & Average $N$ \\ \hline              
      Sunny + Foliage & 3185 & 14361 \\ Sunny + Foliage & 2200 & 17696 \\Cloudy + Foliage & 3312 & 17711  \\Sunny + Foliage & 3620 & 18373 \\Overcast + Mixed Foliage & 3300 & 13770  \\Low Sun + Mixed Foliage & 3286 & 15441  \\Low Sun + Mixed Foliage & 3384 & 16081 \\Cloudy + Mixed Foliage  & 2729 & 14111  \\Low Sun + No Foliage + Snow & 2022 & 19060  \\Overcast + Foliage & 1728 & 20090  \\
      \hline                    
    \end{tabular}
    }%
    \caption{Statistics of the CMU Cross-Season Correspondence dataset. Each row shows the condition, the number of image pairs as well as the mean number of correspondences per image pair for each traversal. Note that there are several traversals with the same condition.}
    \label{tab:cmustat}
\end{table}


\subsection{Oxford RobotCar Correspondence Dataset}
The original RobotCar dataset \cite{maddern20171} was gathered using an autonomous vehicle that traversed a 10 $km$ route in Oxford, UK during 12 months. Similarly as for the CMU dataset, we use the camera poses from ten traversals during different seasons and weather conditions available in the RobotCar Seasons dataset provided by the authors of \cite{sattler2018benchmarking}. For the reference traversal these camera poses were initialized using a GPS/INS system and refined by iteratively triangulating 3D points and performing bundle adjustment. For the other traversals, the camera poses were calculated using 3D points clouds built from the measurements of the LIDAR scanners mounted on the vehicle. The LIDAR point clouds for each traversal were aligned to the LIDAR point cloud of the reference traversal using the Iterative Closest Point algorithm~\cite{Besl1992PAMI} and manual corrections when necessary.

The images included in the RobotCar Seasons dataset are recorded using three synchronized global shutter Point Grey Grasshopper2 cameras mounted to the left, rear, and right of the car. Unfortunately the image quality is poor in general. A lot of the images are overexposed and there is a lot of motion blur present. In addition there is also a lot of image noise for the night time images. This makes the MVS pipeline that we used for the CMU dataset produce point clouds with too few points to be of any use for us. We instead use the LIDAR point clouds available in the original RobotCar dataset. Since we  know the pose of the multi-camera system at each timestamp, as well as the poses of the individual cameras and LIDAR sensors on the car, we can transform the LIDAR point clouds into the coordinate system of the cameras. We then determine which points are visible in which cameras in a separate step. To determine which points are visible in each camera the depth of points projected close to each other in the image is compared.

The matching of the 3D points and pruning of the correspondences are done in the same way as for the CMU Seasons dataset. However, since we do not have as many images for the RobotCar Seasons dataset, we use a larger threshold for the distance between camera pairs, specifically $2.0$ $m$. Table~\ref{tab:oxstat} provides a summary of the content of the final RobotCar Cross-Season Correspondence dataset.

\begin{table}
    \setlength{\tabcolsep}{4pt}
    \centering
    \footnotesize{
    \begin{tabular}{c|c|c}
    Condition & Image pairs & Average $N$ \\ \hline                
      Dawn & 772 & 59158 \\ Dusk & 646 & 50159 \\Night & 646 & 52238  \\Night + Rain & 780 & 43066 \\Summer + Overcast & 809 & 51722  \\Winter + Overcast & 671 & 54466  \\Rain & 683 & 53276 \\Snow  & 823 & 57578  \\Sun & 681 & 48150  \\
      \hline                    
    \end{tabular}
    }%
    \caption{Statistics of the RobotCar Cross-Season Correspondence dataset. Each row shows the condition, the number of image pairs s as well as the mean number of correspondences per image pair for each traversal. Note that there are several traversals with the same condition.}
    \label{tab:oxstat}
\end{table}

\section{Implementation Details}

\begin{figure}
    \centering
    \includegraphics[width=0.23\textwidth]{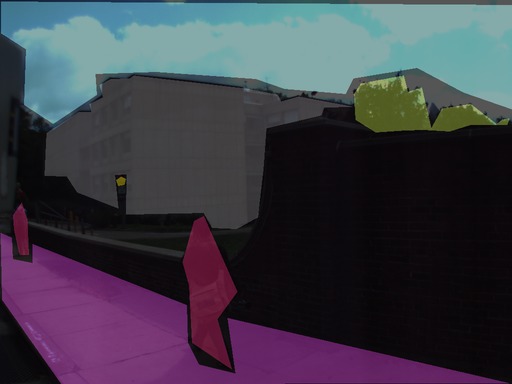}
    \includegraphics[width=0.23\textwidth]{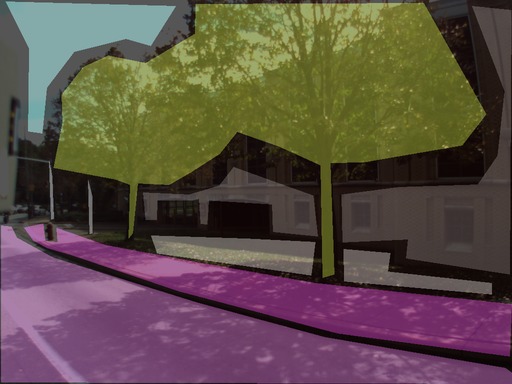}
    \includegraphics[width=0.23\textwidth]{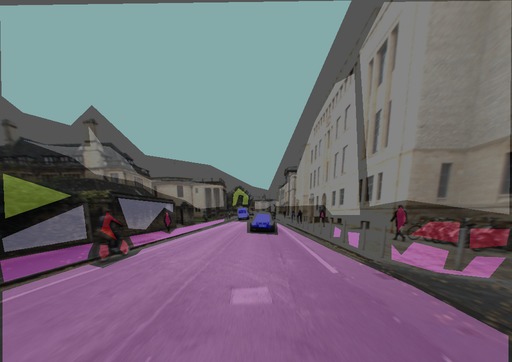}
    \includegraphics[width=0.23\textwidth]{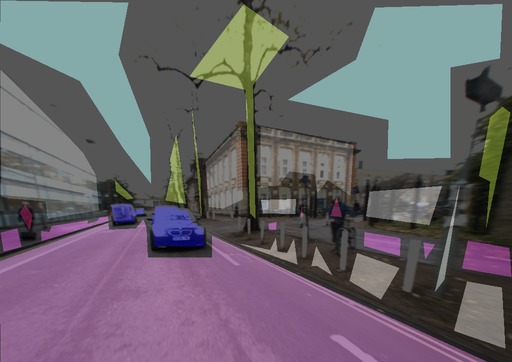}
    \caption{Examples of the manually labeled images added to the training set. The top row shows images for the CMU Seasons dataset (66 images in total) while the bottom row shows images for the RobotCar Seasons dataset (40 images in total).}
    \label{fig:extra_annos}
\end{figure}

During the training of the CNN, we minimize the loss $\mathcal{L}$ described in Section~\ref{sec:loss}. As a starting point, we use the PSPNet \cite{zhao2017pyramid} network pretrained on the Cityscapes dataset \cite{Cordts2016Cityscapes}. In addition to the Cityscapes training images, we also add a few coarsely annotated images from the CMU and RobotCar Seasons datasets, respectively. Some examples of these annotations are shown in  Fig~\ref{fig:extra_annos}. This is necessary to keep the CNN from learning the trivial solution where the same class is predicted for all pixels on the CMU and RobotCar images while still producing good segmentations for the Cityscapes images. Note that only training images with fine annotations from the Cityscapes dataset are used.

We also add an on-the-fly correspondence refinement step, where all correspondences that have their pixel positions in the reference image classified as one of the non-stationary classes are removed. The classes concerned are person, rider, car, truck, bus, train, motorcycle, and bicycle. For a correspondence, if the pixel in the reference image is classified as a non-stationary class it means one of two things: The pixel actually depicts a non-stationary object and  has been incorrectly added to the correspondence dataset, \ie, there is no guarantee that the corresponding pixel in the target image has the same semantic class. The other explanation is that the CNN incorrectly classified the pixel. Adding the loss for said correspondence would not be useful for any of these scenarios. Additionally, we use a warm-up period of 500 iterations before the correspondence loss is added. This ensures that the CNN has produces reasonable segmentations for the reference images.

The optimization method used during training is Stochastic Gradient Descent with momentum and weight decay. During all experiments the learning rate was set to $2.5 \cdot 10^{-5}$, while the momentum and weight decay were set to $0.9$ and $ 10^{-4}$, respectively. In addition, the loss was scaled with $\frac{1}{1 + \lambda}$ to keep the total loss weight to unity. Due to GPU memory limitations we train using a batch size of one. We train the networks for at least 30000 iterations and use the weights that got the best mean Intersection over Union (mIoU) on the validation set. For the RobotCar and CMU Seasons validation and test sets, we take the mean over only classes present in the respective dataset when calculating the mIoU. The number of validation images for the CMU Seasons and Robotcar were 25 and 27 respectively. The corresponding numbers for the test sets were 33 and 27. The training and evaluation is implemented in PyTorch~\cite{paszke2017automatic} and the code is publicly available~\footnote{\url{https://github.com/maunzzz/cross-season-segmentation}}.

All evaluation and testing is done in patches on the original image scale only. The patch size is $713 \times 713$ pixels and the patches are extracted from the image with a step size of $476$ pixels in both directions. The output of the network is paired with an interpolation weight map that is $1$ for the $236 \times 236$ center pixels of the patch and drops off linearly to $0$ at the edges. For each pixel the weighted mean, using the interpolation maps as weights, of all patches that contain it is used to produce the pixel's class scores. The motivation behind the interpolation is that the network generally performs better at the center of the patches, since there is more information about the surroundings available there. 

\section{Experimental Evaluation}
In this section, we present the results of using the Cross-Season Correspondence Dataset to train a CNN for segmentation. The two main points of interest are \textbf{a}), how does the Cross-Season Correspondence Dataset influence the segmentation performance on the images within the dataset, \ie, the CMU and RobotCar images and \textbf{b}), how does using the dataset influence the generalization performance of the segmentor to other datasets. To investigate \textbf{a}), we manually annotate a set of test images from the RobotCar and CMU datasets taken from areas that are not included in the correspondence datasets. 
To answer \textbf{b}), we use the WildDash dataset~\cite{zendel2018wilddash}. The WildDash dataset is designed to evaluate the robustness of segmentation methods and contains many different and challenging images. 

We also investigate the effect of training with correspondences together with different amount of annotated training images. To this end we do experiments both using just the Cityscapes dataset~\cite{Cordts2016Cityscapes} training set as well as, on top of that, adding the training set from the Mapillary Vistas dataset~\cite{neuhold2017mapillary}. The Cityscapes training set contains 2975 annotated images taken under favorable weather conditions and in similar environments while the Vistas training set contains 18000 images from a diverse set of environments, seasons and weather conditions~\cite{neuhold2017mapillary}. As the Vistas dataset contains many more annotated classes than the Cityscapes dataset, we only consider the subset of classes which are in both datasets and treat the rest as unlabeled during training. For all segmentation experiments, we investigate three correspondence loss functions: for the first we use the correspondence cross-entropy loss $l_{CE}$ applied to the final features. For the other two, we apply the correspondence hinge loss $l_{hinge}$ to the final and second to final layer, respectively.



\begin{table}[]
    \centering
    \begin{tabular}{|c?c|c|c|}
        \hline
          $\lambda$ & CE & Hinge$_C$ & Hinge$_F$ \\ 
         \hline 
         $0.01$ & $63.3$ & $62.0 $ & $62.2 $\\
         $0.05$ & $64.4$ & $62.6$ &  $62.1$\\
         $0.1$ & $65.6$ & $63.2$ & $\bm{62.3}$ \\
         $0.5$ & $65.0$ & $62.3$  & $61.9 $ \\
         $1.0$ & $\bm{67.2}$ & $\bm{64.0}$  & $60.6 $ \\
         $2.0$ & $66.0$ & $63.3$  & $59.5 $ \\
         $5.0$ & $66.7$ & $62.6$  & $59.4 $ \\
         \hline
    \end{tabular}
    \caption{Parameter study of $\lambda$ for the loss $\mathcal{L} = \mathcal{L}_{sup} + \lambda \mathcal{L}_{corr}$ (\cf Section~\ref{sec:loss}) on the CMU dataset.
    The correspondence losses included are cross-entropy loss $l_{CE}$ and hinge loss $l_{hinge}$ applied to the final and second to final features, Hinge$_C$ and Hinge$_F$, respectively. All values are mIoU (in $\%$) for the CMU validation set. }
    \label{tab:cmu_par}
\end{table}

\PAR{Parameter Study. }
The parameter $\lambda$ specifies the trade-off between the fully supervised cross-entropy loss ($\mathcal{L}_{sup}$) for the annotated training set and the correspondence loss ($\mathcal{L}_{corr}$). A higher $\lambda$ means that more emphasis is put on minimizing the correspondence loss compared to the fully supervised loss. To investigate the impact of $\lambda$, we perform a parameter study which is summarized in Table~\ref{tab:cmu_par}. The results presented are the mIoU on the CMU validation set. As can be seen from the table, the best choice of $\lambda$ for the correspondence cross-entropy loss (CE) and the hinge loss on the final features (Hinge$_C$) is $\lambda = 1.0$ while for the hinge loss applied to the second to final feature layers (Hinge$_F$) it is $\lambda = 0.1$. We hence choose these values for the remaining experiments. 

\begin{table}
\begin{center}
\begin{tabular}{c|cc?cc|cc|}
\cline{2-7}
& \multirow{2}{*}{Extra} & \multirow{2}{*}{Corr} & \multicolumn{2}{c|}{CMU} & \multicolumn{2}{c|}{RobotCar} \\ \cline{4-7}
 &  &  & CMU & WD & RC & WD \\ \cline{2-7}
\multirow{5}{*}{\begin{turn}{90} CS \end{turn}}    & & & $31.2$ & $16.4 $ & $22.2  $ & $16.4 $\\
& \checkmark &  & $73.6  $ & $37.0 $ &  $45.8  $ & $25.4 $\\
& \checkmark & CE & $\bm{79.3} $ & $\bm{39.6}  $ & $53.8 $ & $27.8$\\
& \checkmark & Hinge$_C$ & $72.4 $ & $37.9  $ & $50.6 $ & $25.2  $\\
& \checkmark & Hinge$_F$ & $75.3 $ & $38.7 $ & $ \bm{55.4} $ & $\bm{27.9}  $\\ \cline{2-7}
\multirow{5}{*}{\begin{turn}{90}CS + Vistas\end{turn}} & &  &  $77.4 $ & $49.3  $ & $46.8 $ & $\bm{49.2}  $\\
  & \checkmark &  & $82.8  $ & $51.5 $  & $51.2  $ & $46.4 $\\
 & \checkmark & CE & $\bm{85.9} $ & $52.9  $  & $\bm{59.4}  $ & $48.5 $ \\
 & \checkmark & Hinge$_C$ & $82.2 $ & $\bm{54.9} $ & $59.0  $ & $47.9 $ \\
 & \checkmark & Hinge$_F$ & $84.0 $ & $54.5  $  & $58.7 $ & $45.6  $\\ \cline{2-7}
\end{tabular}
\end{center}
\caption{\label{res:both} Segmentation results for the models trained on the CMU correspondence data (left) and the RobotCar correspondence data (right). Results for the CMU test set, the RobotCar (RC) test set and the WildDash (WD) validation set are shown in terms of mIoU $\%$. For the bottom five rows the Vistas training set was used in addition to Cityscapes (CS). 
Column one marks if the extra training annotations from the CMU/RobotCar dataset were used. Column two specifies the correspondence training loss used, \ie, correspondence cross-entropy loss (CE) applied to the final layer and hinge loss applied to the final and second to final features, Hinge$_C$ and Hinge$_F$, respectively.}
\end{table}

\def \imwww {0.16}
\newcommand{\cmuima}{img_03803_c0_1287504141845554us}
\newcommand{\cmuimb}{img_05048_c0_1288106678545243us}
\newcommand{\cmuimc}{img_05160_c0_1290445101652483us}
\newcommand{\cmuimd}{img_05240_c1_1289590095130052us}
\newcommand{\cmuime}{img_05372_c0_1284563720487997us}
\newcommand{\cmuimf}{img_05378_c0_1303398948248624us}
\newcommand{\cmuimg}{img_05533_c0_1292959419262850us}
\newcommand{\cmuimh}{img_05653_c1_1284563744683822us}
\newcommand{\cmuimi}{img_05735_c0_1290445161174594us}
\newcommand{\cmuimj}{img_05959_c1_1292959456263219us}
\newcommand{\cmuimk}{img_07019_c1_1283348377844918us}
\begin{figure*}
    \centering
    \setlength\tabcolsep{1pt} 
    \begin{tabular}{cccccc}
        Image & Annotation & E & E + C & V + E & V + E + C \\
        \includegraphics[width=\imwww\textwidth]{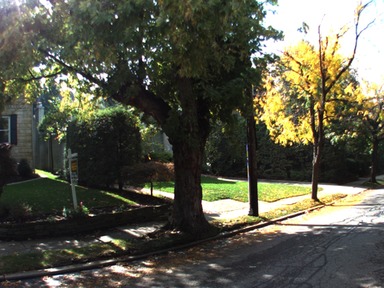} &
        \includegraphics[width=\imwww\textwidth]{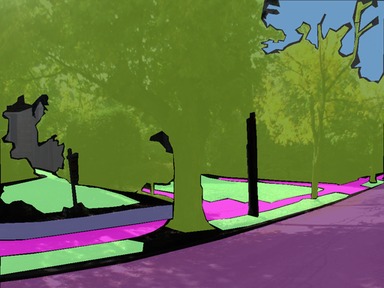} &
        \includegraphics[width=\imwww\textwidth]{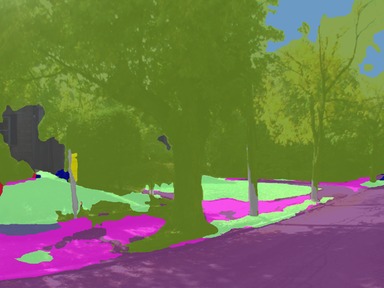} &
        \includegraphics[width=\imwww\textwidth]{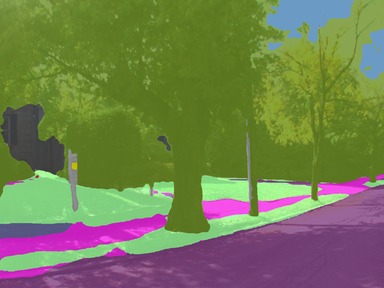} &
        \includegraphics[width=\imwww\textwidth]{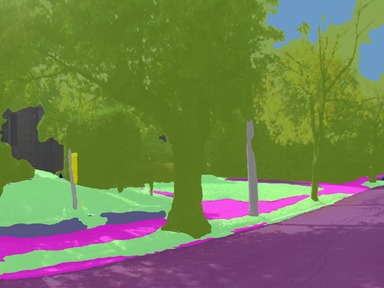} &
        \includegraphics[width=\imwww\textwidth]{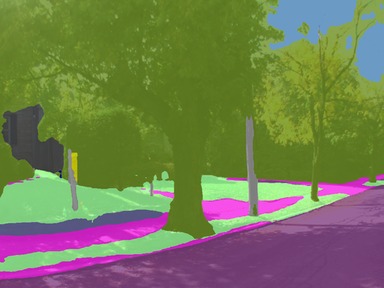} \\

        \includegraphics[width=\imwww\textwidth]{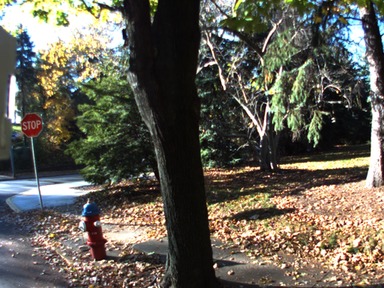} &
        \includegraphics[width=\imwww\textwidth]{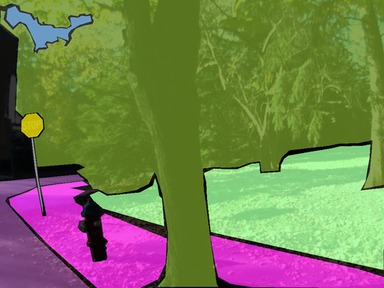} &
        \includegraphics[width=\imwww\textwidth]{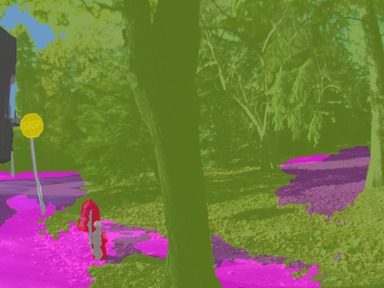} &
        \includegraphics[width=\imwww\textwidth]{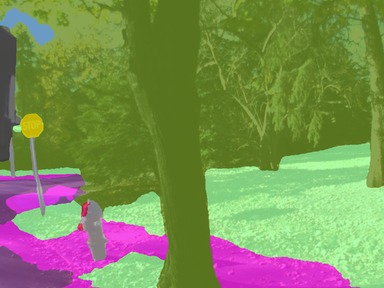} &
        \includegraphics[width=\imwww\textwidth]{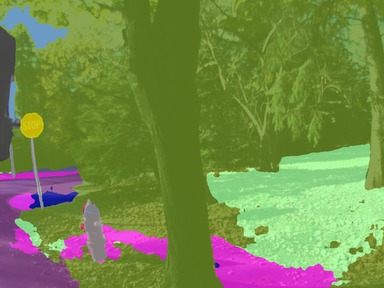} &
        \includegraphics[width=\imwww\textwidth]{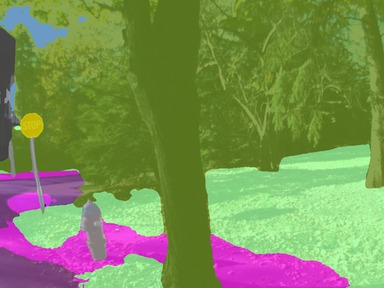} \\

        \includegraphics[width=\imwww\textwidth]{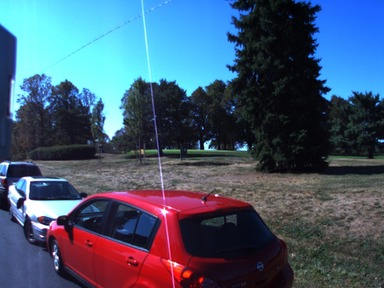} &
        \includegraphics[width=\imwww\textwidth]{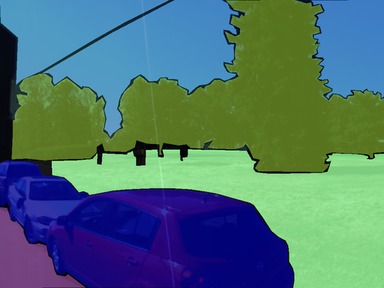} &
        \includegraphics[width=\imwww\textwidth]{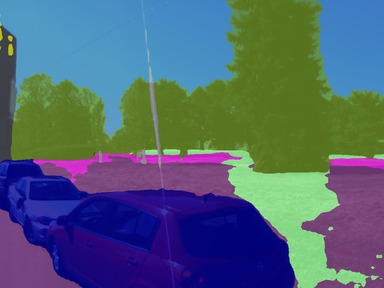} &
        \includegraphics[width=\imwww\textwidth]{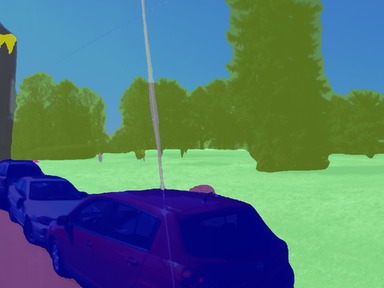} &
        \includegraphics[width=\imwww\textwidth]{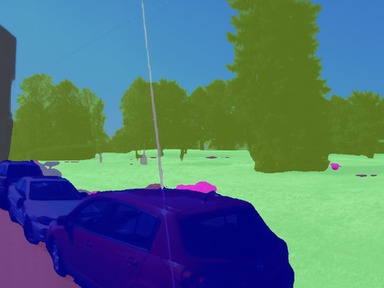} &
        \includegraphics[width=\imwww\textwidth]{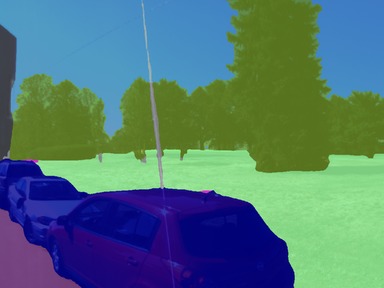} \\

        \includegraphics[width=\imwww\textwidth]{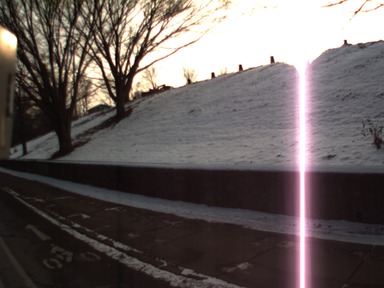} &
        \includegraphics[width=\imwww\textwidth]{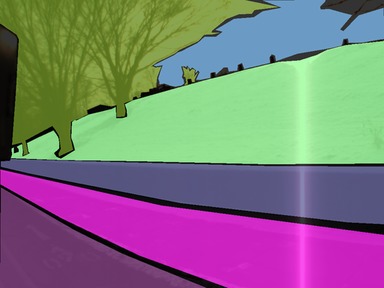} &
        \includegraphics[width=\imwww\textwidth]{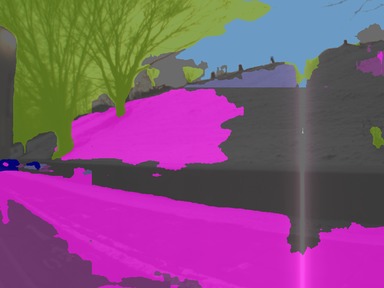} &
        \includegraphics[width=\imwww\textwidth]{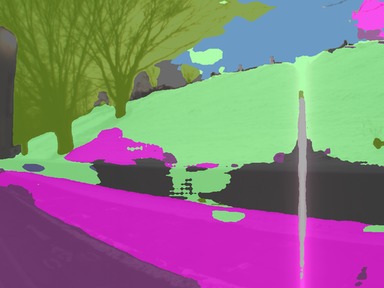} &
        \includegraphics[width=\imwww\textwidth]{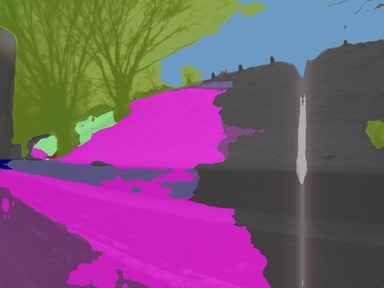} &
        \includegraphics[width=\imwww\textwidth]{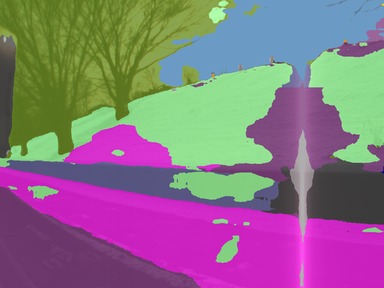} \\

    \end{tabular}
    \setlength\tabcolsep{6pt} 
    \caption{Qualitative results on the CMU test set. Four different networks are compared, the notation used is: E: trained with extra CMU annotations, C: trained with correspondence data, V: trained with Vistas training set. The most notable performance difference when adding correspondences are for areas that are visually different between seasons. This can be seen for the terrain areas covered in leaves for row two and three as well as the area of snow in row four. The image in row four is especially challenging since it contains a lot of snow as well as an apparent lens glare. Yet, the networks trained with correspondences still manages to get part of the snow patch correctly labeled.}
    \label{fig:cmu_ims}
\end{figure*}

\def \imwww {0.16}
\newcommand{\oxima}{dawnleft1418723512039667}
\newcommand{\oximb}{night-rainrear1418841246916088}
\newcommand{\oximc}{night-rainright1418841082436378}
\newcommand{\oximd}{nightrear1418236099050440}
\newcommand{\oxime}{overcast-summerrear1432294544193098}
\newcommand{\oximf}{rainrear1416908256629467}
\begin{figure*}
    \centering
    \setlength\tabcolsep{1pt} 
    \begin{tabular}{cccccc}
        Image & Annotation & E & E + C & V + E & V + E + C \\

        \includegraphics[width=\imwww\textwidth]{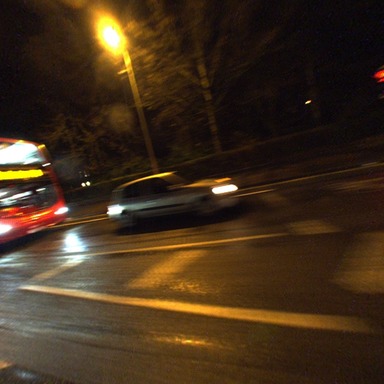} &
        \includegraphics[width=\imwww\textwidth]{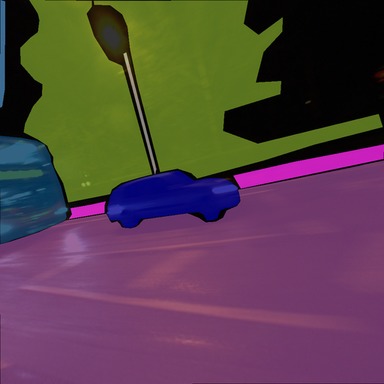} &
        \includegraphics[width=\imwww\textwidth]{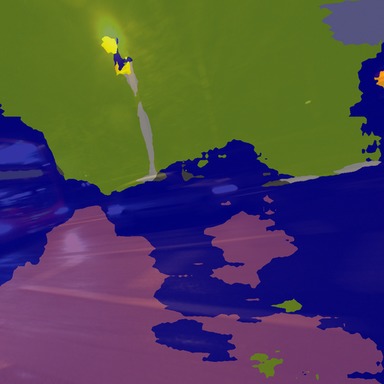} &
        \includegraphics[width=\imwww\textwidth]{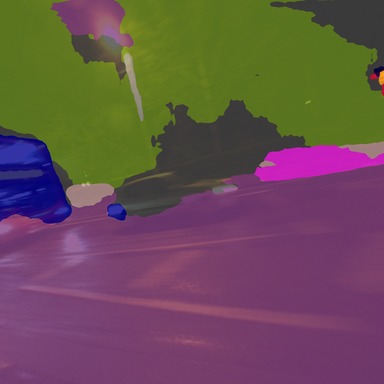} &
        \includegraphics[width=\imwww\textwidth]{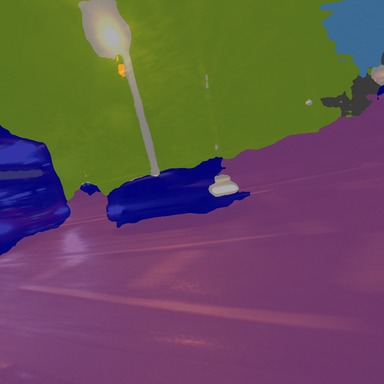} &
        \includegraphics[width=\imwww\textwidth]{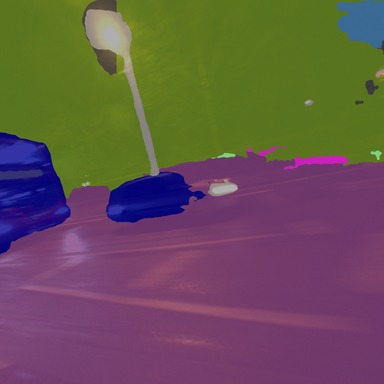} \\
        
        \includegraphics[width=\imwww\textwidth]{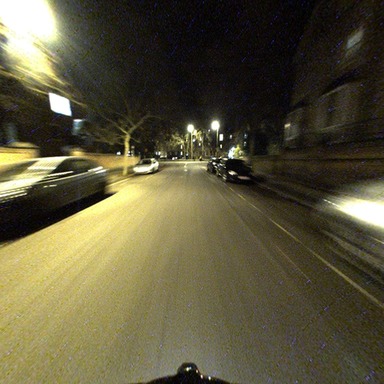} &
        \includegraphics[width=\imwww\textwidth]{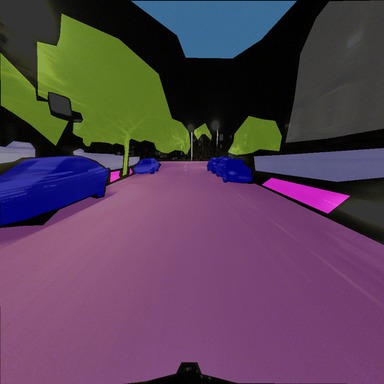} &
        \includegraphics[width=\imwww\textwidth]{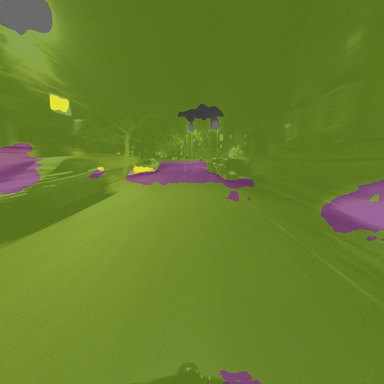} &
        \includegraphics[width=\imwww\textwidth]{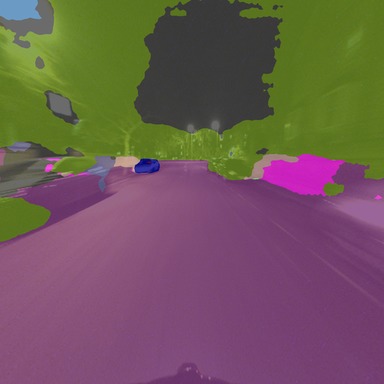} &
        \includegraphics[width=\imwww\textwidth]{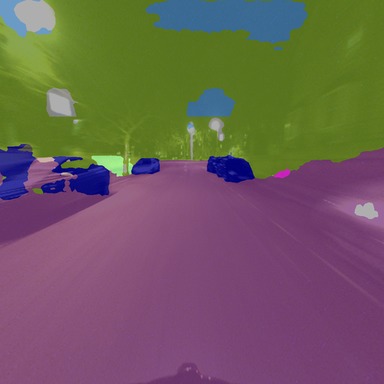} &
        \includegraphics[width=\imwww\textwidth]{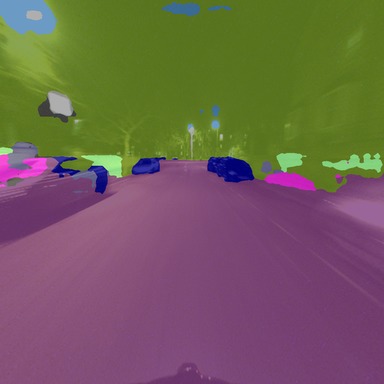} \\

        \includegraphics[width=\imwww\textwidth]{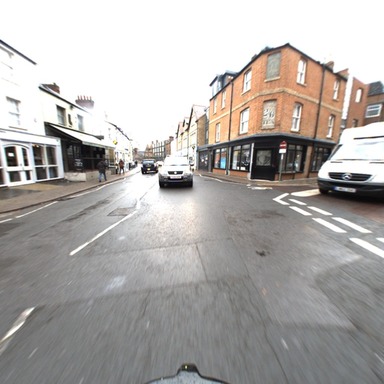} &
        \includegraphics[width=\imwww\textwidth]{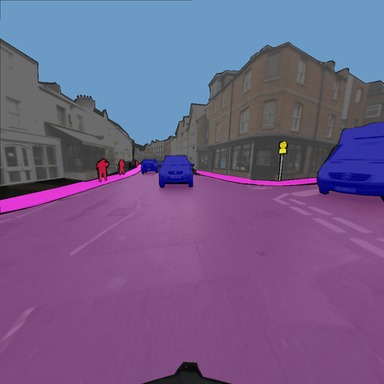} &
        \includegraphics[width=\imwww\textwidth]{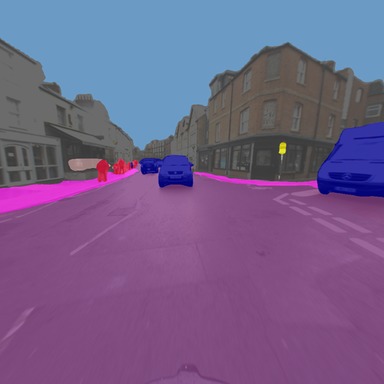} &
        \includegraphics[width=\imwww\textwidth]{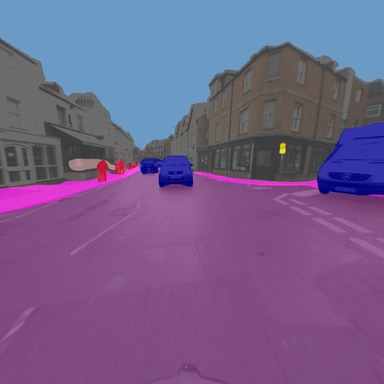} &
        \includegraphics[width=\imwww\textwidth]{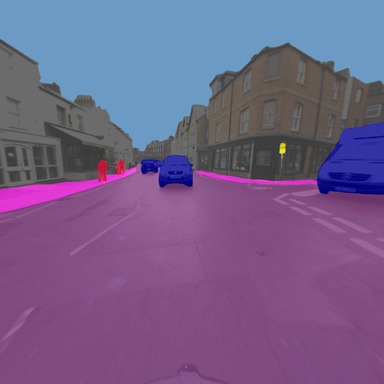} &
        \includegraphics[width=\imwww\textwidth]{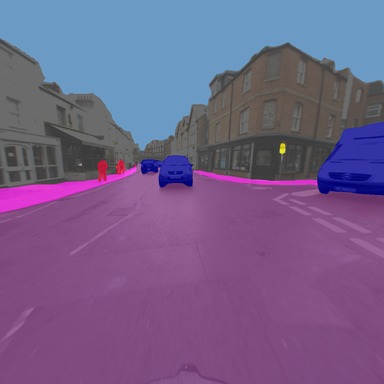} \\
    \end{tabular}
    \setlength\tabcolsep{6pt} 
    \caption{Qualitative results on the RobotCar test set. Four different networks are compared, the notation used is: E: trained with extra RobotCar annotations, C: trained with correspondence data, V: trained with Vistas training set. The most notable performance difference when adding correspondences are for areas that are on the night images, row one and two. Comparing the results of E and E + C, we can see that adding correspondences enables the network to label the road correctly. It however fails at labeling sky and cars correctly since these are not included in the correspondence dataset.}
    \label{fig:ox_ims}
\end{figure*}

\PAR{Segmentation.}
Table~\ref{res:both} summarizes the results of the segmentation experiments on the CMU as well as the RobotCar datasets. For the networks trained on the CMU dataset, the mIoU on the CMU test set and the WildDash validation set is presented for several baselines as well as networks trained with the correspondence dataset. For the networks trained on the RobotCar dataset the RobotCar test set is used instead of the CMU test set. Comparing the two baselines, there is a large difference in performance between the network trained with both Cityscapes and Vistas compared to the one trained with Cityscapes only. This is to be expected since the Vistas dataset has much more diversity when it comes to seasons and weather conditions, enabling a network trained on it to generalize well to the CMU or RobotCar test set. 

When investigating the effect of the correspondence dataset, it is reasonable to compare the network trained with the extra annotations with the networks trained with extra annotations and the correspondence loss. As can be seen in Table~\ref{res:both}, we get an improvement in mIoU when adding the correspondence loss for both the network trained with Cityscapes and extra annotations as well as the one trained with Cityscapes, Vistas and extra annotations. This holds both for the CMU and RobotCar datset. As can be seen in Fig.~\ref{fig:cmu_ims} where qualitative results on the CMU test set are presented, adding the correspondences improves segmentations for areas that are very visually different between seasons. Some examples are yellow leaves on the ground during autumn or snow during the winter. These confuse the network trained only on Cityscapes since there are no such examples in the training set. The baseline that is also trained on Vistas handles these situations better but there is still an improvement when adding the correspondence training, especially when it comes to, \eg, patches covered in snow, as these are not included in the subset of classes present in both datasets and, thus, excluded from the training.  For the RobotCar dataset, the most prevalent improvements are on the night images, which can be seen in some example segmentations in Fig.~\ref{fig:ox_ims}. Additional qualitative results can be seen in the supplementary material. Worth to note is also that we manage to get a larger performance increase using just a few coarsely annotated images and the correspondence dataset compared to adding the entire Vistas training set, both for the CMU and the RobotCar data. The Cross-Season Correspondence datasets required about 30 hours of manual labour each in the form of annotating correspondences and verifying poses. Adding the two hours it took to coarsely annotate a few images gives a total of 32 hours of manual labour required. This, compared to an estimated 28200 hours to annotate the Vistas training set.  

The best performing correspondence loss differs between the datasets. On the CMU dataset, the CE loss performs best both with and without the Vistas training set. However, for the RobotCar dataset without the Vistas labels, the loss that performs best is Hinge$_F$. A difference between Hinge$_F$ and CE is that CE imposes a harder constraint on the output of the network for the correspondences. It basically treats the output of the network on the reference image as the ground truth for the corresponding pixels in the target image. Since the RobotCar Cross-Season Correspondence dataset is created using LIDAR data, the point measurements are not perfectly synchronized with the images which creates a slight misalignment for some correspondences. If there are some erroneous correspondences, having a hard constraint can be harmful for the performance of the network. In these cases, imposing a softer constraint via the Hinge$_F$ loss, which basically stipulates that the features should be similar, can give a better performance, especially when using a strong baseline training set. 

Adding the correspondence training does improve the segmentation performance on the WildDash images slightly for the CMU dataset but not for the RobotCar dataset. A reason that there is no significant performance gain could be that the camera poses relative to the vehicle (front facing) and the image resolution are very similar for the Cityscapes, Vistas and WildDash datasets. For the CMU Seasons dataset, there is no front facing camera, only one facing forward/left and one forward/right. Hence, learning to segment these well does not necessarily improve the segmentation performance on WildDash. In addition, the image quality for both the CMU and the RobotCar images are poorer than those of Cityscapes, Vistas and WildDash. Despite this, adding the correspondence loss improves the results for the network trained on the CMU data and on the RobotCar data without Vistas. The advantage of learning to segment images during other weather conditions seems to be large enough to make a difference on the WildDash validation set.

\section{Conclusion}
In this paper, we have introduced two Cross-Season Correspondence datasets, each consisting of a set of 2D-2D matches between images taken under different conditions. We described how these datasets can be generated with little human supervision and demonstrated the usefulness of the datasets by training an image segmentation network. To this end we presented and investigated three different training losses, based on cross-entropy and hinge loss, that can be used for the correspondence data. Our experiments showed that adding the correspondences as extra supervision during training improves the segmentation performance of the network, making it more robust to seasonal changes and weather conditions. Improving the semantic segmentation performance could in turn lead to more robust localization results which provides the first step towards an iterative feedback loop improving localization and semantic segmentation. An investigation on how the improved image segmentation network affect semantic localization methods is left for future work.


Important future research directions include investigating options to remove the need for a few, manually annotated images. Possible approaches include an additional unsupervised domain adaptation step, adapting the segmentation algorithm to the reference images of the correspondence dataset. Additionally, the Cross-Season Correspondence Datasets provide opportunities for other application such as training robust feature detectors and descriptors.



{\footnotesize \PAR{Acknowledgements} This work has been funded by the Swedish Research Council (grant no. 2016-04445), the Swedish Foundation for Strategic Research (Semantic Mapping and Visual Navigation for Smart Robots) and Vinnova / FFI (Perceptron, grant no. 2017-01942).}

\clearpage
\appendix
\section*{Supplementary Material}

This supplementary material shows additional qualitative results for the CMU Seasons~\cite{badino2011visual,sattler2018benchmarking} and RobotCar Seasons~\cite{maddern20171,sattler2018benchmarking} datasets. These results were not included in the main paper due to space constraints. 

We show additional qualitative results for the CMU Seasons~\cite{badino2011visual,sattler2018benchmarking} dataset in Fig.~\ref{fig:cmu_ims_supp} and Fig.~\ref{fig:cmu_ims_supp_vis}. Fig.~\ref{fig:cmu_ims_supp} shows results on the test set used to measure the segmentation quality quantitatively via the mean IoU score. Consequently, we also show the reference annotations. 
Fig.~\ref{fig:cmu_ims_supp_vis} shows additional results on unannotated images from the CMU Seasons dataset. 
As can be expected, using correspondences, as proposed in the paper, mainly improves segmentation quality in areas most affected by seasonal changes, \eg, roads, side walks, and terrain areas covered in leaves or snow.

Similarly, additional example segmentations for the RobotCar~\cite{maddern20171,sattler2018benchmarking} dataset can be seen in Fig.~\ref{fig:ox_ims_supp} and Fig.~\ref{fig:ox_ims_supp_vis}. In addition to improving the segmentation performance on the night images, adding correspondences helps with segmentation of the overexposed parts of buildings, see for example row one of Fig.~\ref{fig:ox_ims_supp_vis}. 

For both datasets, we see a clear improvement in segmentation quality when using correspondences (E + C) compared to only using Cityscapes~\cite{Cordts2016Cityscapes} and annotated dataset images (E) for training. 
This is due to the fact that the Cityscapes dataset does not exhibit strong seasonal or illumination changes. 
In contrast, the Mapillary Vistas dataset~\cite{neuhold2017mapillary} contains images captured under a much more diverse set of conditions. 
Still, we observe an improvement in segmentation quality when using correspondences (V + E + C) compared to not using them (V + E).

\def \imwww {0.16}
\newcommand{\supcmuima}{img_04150_c1_1287504175706125us}
\newcommand{\supcmuimb}{img_05048_c0_1288106678545243us}
\newcommand{\supcmuimc}{img_05160_c0_1290445101652483us}
\newcommand{\supcmuimd}{img_05240_c1_1289590095130052us}
\newcommand{\supcmuime}{img_05372_c0_1284563720487997us}
\newcommand{\supcmuimf}{img_05378_c0_1303398948248624us}
\newcommand{\supcmuimg}{img_05533_c0_1292959419262850us}
\newcommand{\supcmuimh}{img_05653_c1_1284563744683822us}
\newcommand{\supcmuimi}{img_05735_c0_1290445161174594us}
\newcommand{\supcmuimj}{img_05959_c1_1292959456263219us}
\newcommand{\supcmuimk}{img_07019_c1_1283348377844918us}
\begin{figure*}
    \centering
    \setlength\tabcolsep{1pt} 
    \begin{tabular}{cccccc}
        Image & Annotation & E & E + C & V + E & V + E + C \\
        \includegraphics[width=\imwww\textwidth]{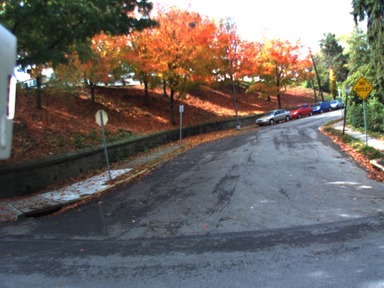} &
        \includegraphics[width=\imwww\textwidth]{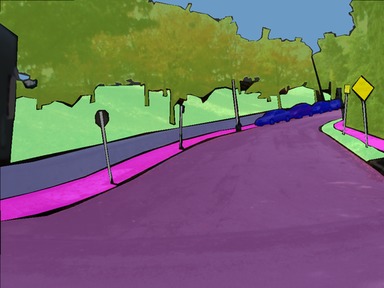} &
        \includegraphics[width=\imwww\textwidth]{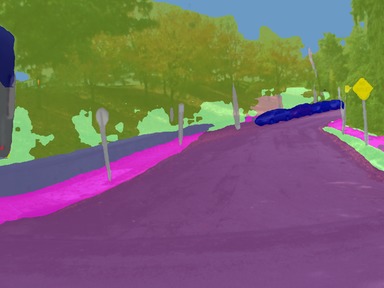} &
        \includegraphics[width=\imwww\textwidth]{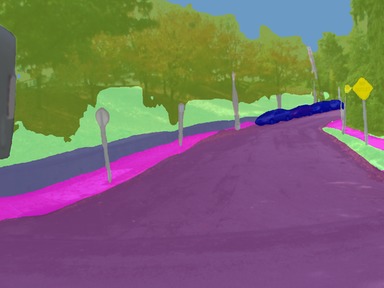} &
        \includegraphics[width=\imwww\textwidth]{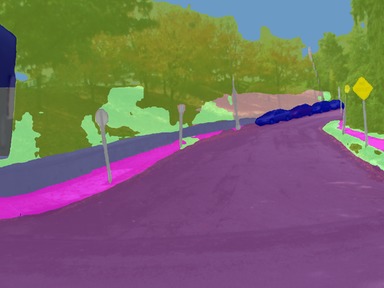} &
        \includegraphics[width=\imwww\textwidth]{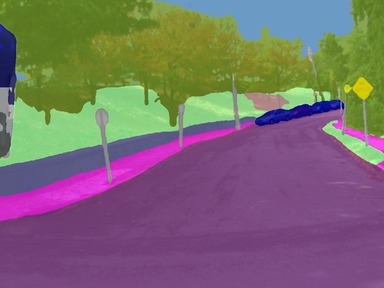} \\
        
        \includegraphics[width=\imwww\textwidth]{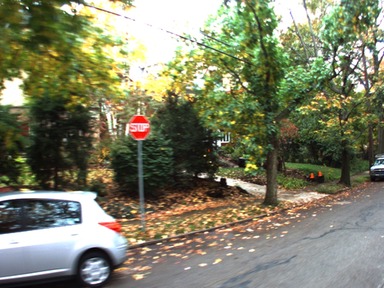} &
        \includegraphics[width=\imwww\textwidth]{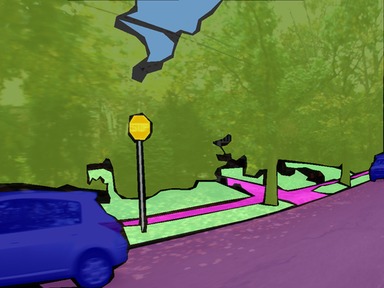} &
        \includegraphics[width=\imwww\textwidth]{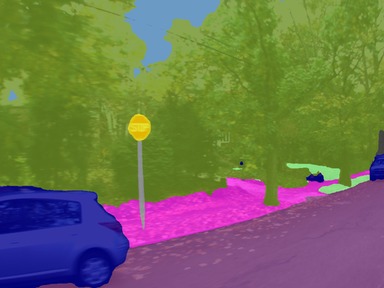} &
        \includegraphics[width=\imwww\textwidth]{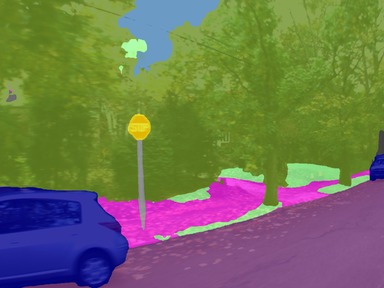} &
        \includegraphics[width=\imwww\textwidth]{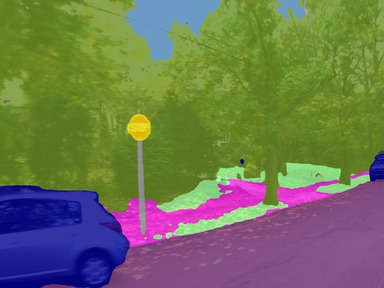} &
        \includegraphics[width=\imwww\textwidth]{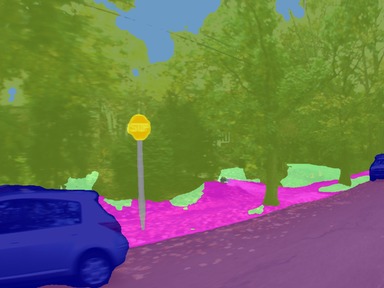} \\
        
        \includegraphics[width=\imwww\textwidth]{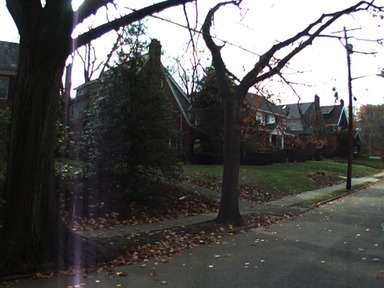} &
        \includegraphics[width=\imwww\textwidth]{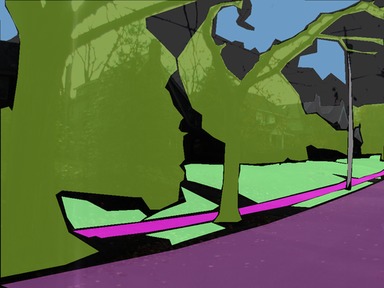} &
        \includegraphics[width=\imwww\textwidth]{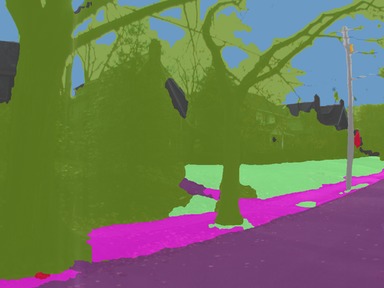} &
        \includegraphics[width=\imwww\textwidth]{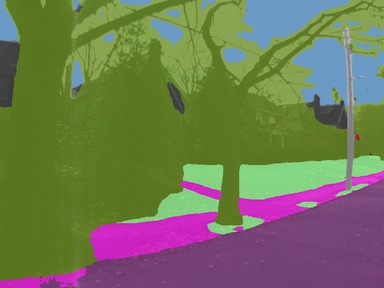} &
        \includegraphics[width=\imwww\textwidth]{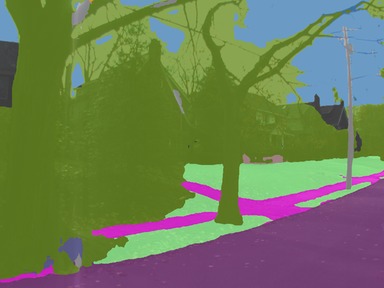} &
        \includegraphics[width=\imwww\textwidth]{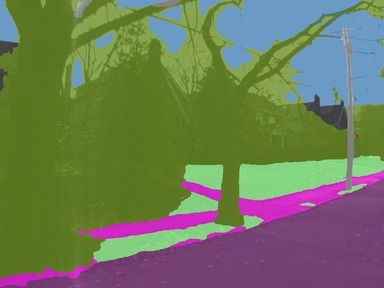} \\
        
        \includegraphics[width=\imwww\textwidth]{figures/results/cmu_test_res/\supcmuimd.jpg} &
        \includegraphics[width=\imwww\textwidth]{figures/results/cmu_test_res/\supcmuimd_truth.jpg} &
        \includegraphics[width=\imwww\textwidth]{figures/results/cmu_test_res/\supcmuimd_cs-extra.jpg} &
        \includegraphics[width=\imwww\textwidth]{figures/results/cmu_test_res/\supcmuimd_cs-extra-corr.jpg} &
        \includegraphics[width=\imwww\textwidth]{figures/results/cmu_test_res/\supcmuimd_cs-map-extra.jpg} &
        \includegraphics[width=\imwww\textwidth]{figures/results/cmu_test_res/\supcmuimd_cs-map-extra-corr.jpg} \\
        
        \includegraphics[width=\imwww\textwidth]{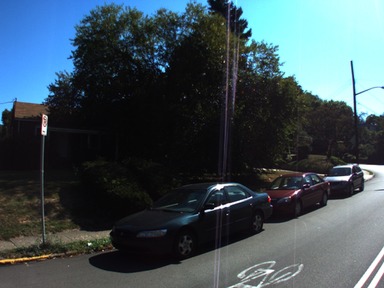} &
        \includegraphics[width=\imwww\textwidth]{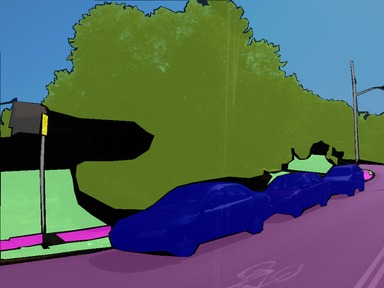} &
        \includegraphics[width=\imwww\textwidth]{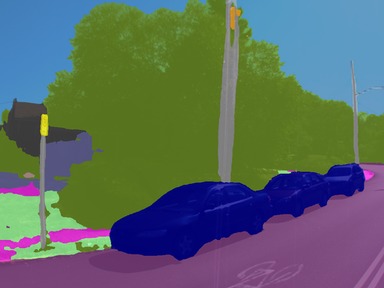} &
        \includegraphics[width=\imwww\textwidth]{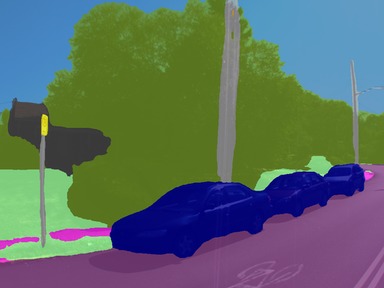} &
        \includegraphics[width=\imwww\textwidth]{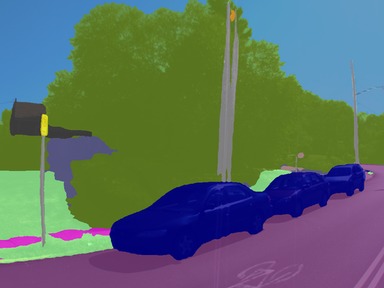} &
        \includegraphics[width=\imwww\textwidth]{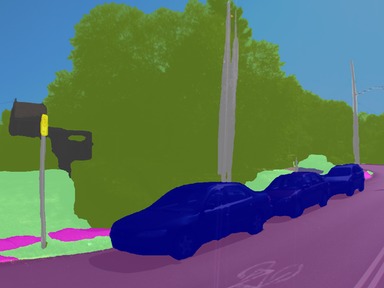} \\
        
        \includegraphics[width=\imwww\textwidth]{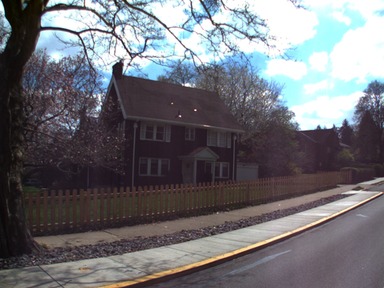} &
        \includegraphics[width=\imwww\textwidth]{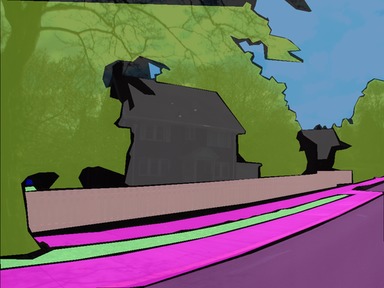} &
        \includegraphics[width=\imwww\textwidth]{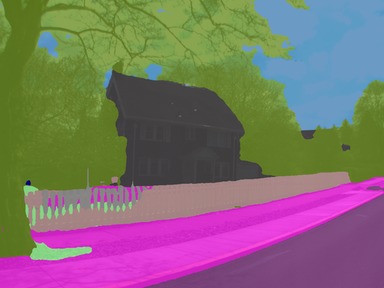} &
        \includegraphics[width=\imwww\textwidth]{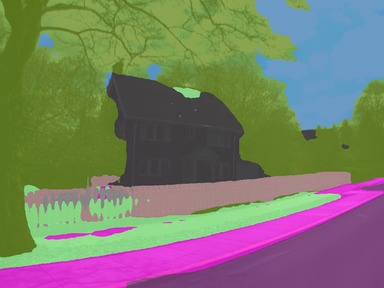} &
        \includegraphics[width=\imwww\textwidth]{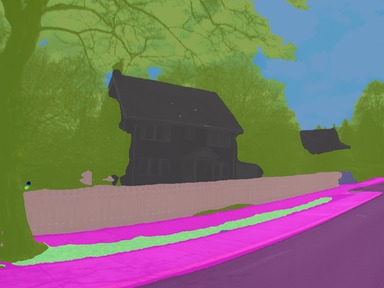} &
        \includegraphics[width=\imwww\textwidth]{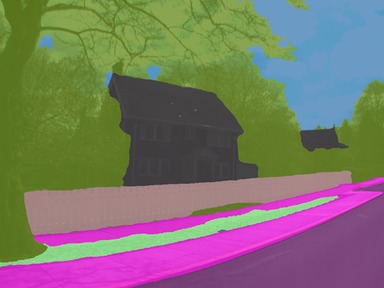} \\

        \includegraphics[width=\imwww\textwidth]{figures/results/cmu_test_res/\supcmuimh.jpg} &
        \includegraphics[width=\imwww\textwidth]{figures/results/cmu_test_res/\supcmuimh_truth.jpg} &
        \includegraphics[width=\imwww\textwidth]{figures/results/cmu_test_res/\supcmuimh_cs-extra.jpg} &
        \includegraphics[width=\imwww\textwidth]{figures/results/cmu_test_res/\supcmuimh_cs-extra-corr.jpg} &
        \includegraphics[width=\imwww\textwidth]{figures/results/cmu_test_res/\supcmuimh_cs-map-extra.jpg} &
        \includegraphics[width=\imwww\textwidth]{figures/results/cmu_test_res/\supcmuimh_cs-map-extra-corr.jpg} \\

        \includegraphics[width=\imwww\textwidth]{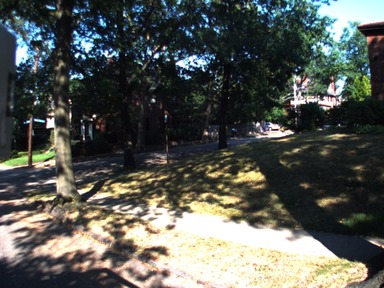} &
        \includegraphics[width=\imwww\textwidth]{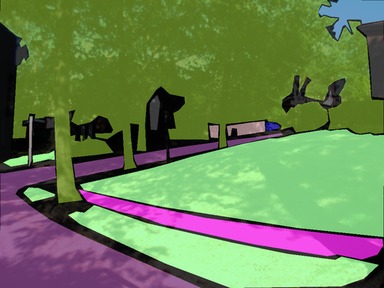} &
        \includegraphics[width=\imwww\textwidth]{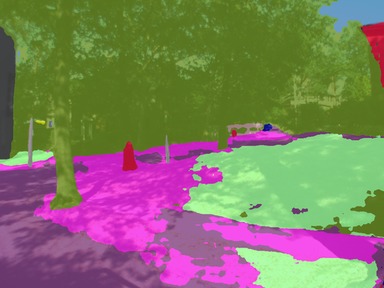} &
        \includegraphics[width=\imwww\textwidth]{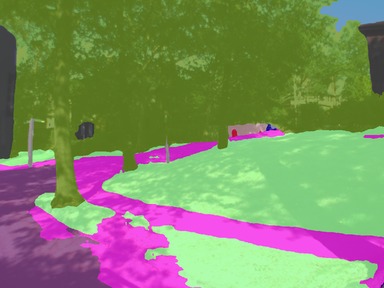} &
        \includegraphics[width=\imwww\textwidth]{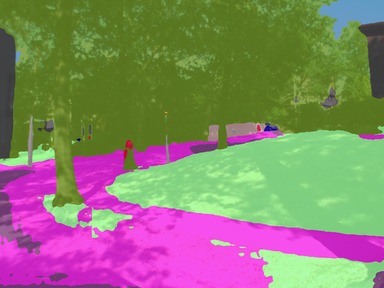} &
        \includegraphics[width=\imwww\textwidth]{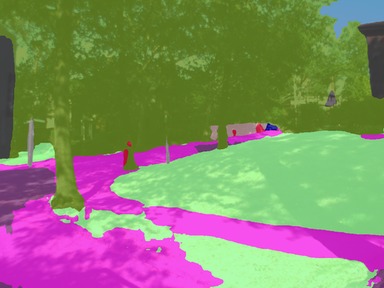} \\
        
    \end{tabular}
    \setlength\tabcolsep{6pt} 
    \caption{Qualitative results on the CMU Seasons test set. Four different networks are compared, the notations used are: E: trained with extra CMU annotations, C: trained with correspondence data, V: trained with Vistas training set. Row two shows a failure case where V + E + C miss-labels terrain as sidewalk.}
    \label{fig:cmu_ims_supp}
\end{figure*}

\def \imwww {0.19}
\newcommand{\viscmuima}{img_04564_c0_1287504213365904us}
\newcommand{\viscmuimb}{img_04806_c0_1292959356328666us}
\newcommand{\viscmuimc}{img_05373_c0_1299251005913800us}
\newcommand{\viscmuimd}{img_05419_c1_1290445128580747us}
\newcommand{\viscmuime}{img_05620_c0_1292959426529568us}
\newcommand{\viscmuimf}{img_06053_c1_1288792854621092us}
\newcommand{\viscmuimg}{img_06926_c1_1299251142514568us}
\newcommand{\viscmuimh}{img_07262_c1_1299251172581504us}
\newcommand{\viscmuimi}{img_07361_c1_1288106889640693us}
\newcommand{\viscmuimj}{img_07948_c0_1284563947448113us}
\newcommand{\viscmuimk}{img_08116_c1_1292959643598564us}
\newcommand{\viscmuiml}{img_10770_c1_1283348710786397us}
\begin{figure*}
    \centering
    \setlength\tabcolsep{1pt} 
    \begin{tabular}{ccccc}
        Image & E & E + C & V + E & V + E + C \\
        \includegraphics[width=\imwww\textwidth]{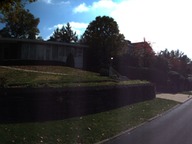} &
        \includegraphics[width=\imwww\textwidth]{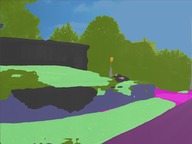} &
        \includegraphics[width=\imwww\textwidth]{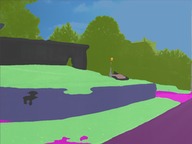} &
        \includegraphics[width=\imwww\textwidth]{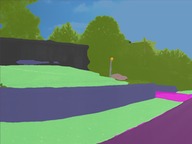} &
        \includegraphics[width=\imwww\textwidth]{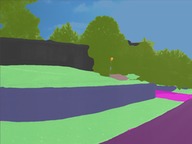} \\

        \includegraphics[width=\imwww\textwidth]{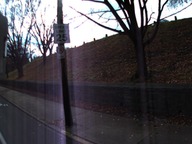} &
        \includegraphics[width=\imwww\textwidth]{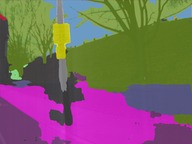} &
        \includegraphics[width=\imwww\textwidth]{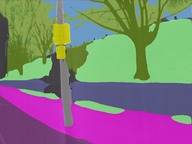} &
        \includegraphics[width=\imwww\textwidth]{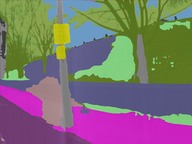} &
        \includegraphics[width=\imwww\textwidth]{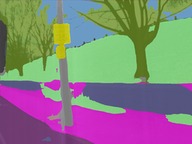} \\
        
        \includegraphics[width=\imwww\textwidth]{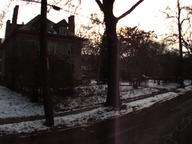} &
        \includegraphics[width=\imwww\textwidth]{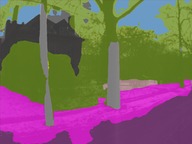} &
        \includegraphics[width=\imwww\textwidth]{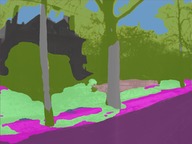} &
        \includegraphics[width=\imwww\textwidth]{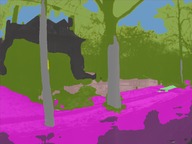} &
        \includegraphics[width=\imwww\textwidth]{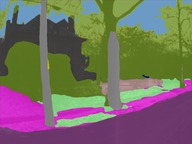} \\

        \includegraphics[width=\imwww\textwidth]{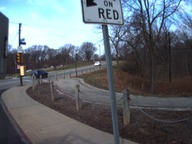} &
        \includegraphics[width=\imwww\textwidth]{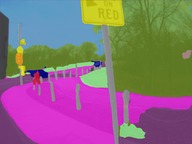} &
        \includegraphics[width=\imwww\textwidth]{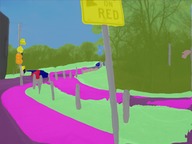} &
        \includegraphics[width=\imwww\textwidth]{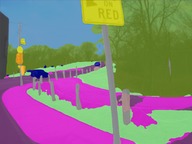} &
        \includegraphics[width=\imwww\textwidth]{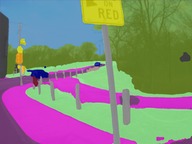} \\
        
        \includegraphics[width=\imwww\textwidth]{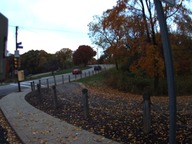} &
        \includegraphics[width=\imwww\textwidth]{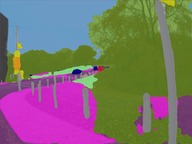} &
        \includegraphics[width=\imwww\textwidth]{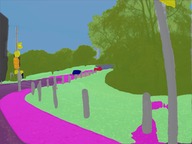} &
        \includegraphics[width=\imwww\textwidth]{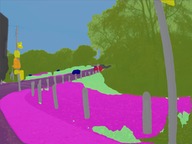} &
        \includegraphics[width=\imwww\textwidth]{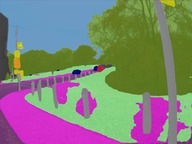} \\
        
        \includegraphics[width=\imwww\textwidth]{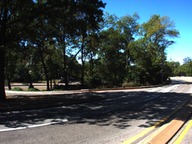} &
        \includegraphics[width=\imwww\textwidth]{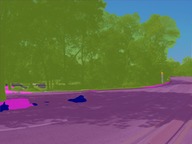} &
        \includegraphics[width=\imwww\textwidth]{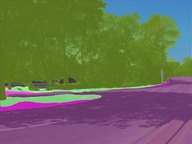} &
        \includegraphics[width=\imwww\textwidth]{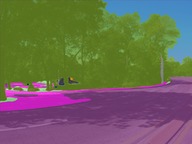} &
        \includegraphics[width=\imwww\textwidth]{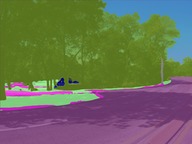} \\
        
        \includegraphics[width=\imwww\textwidth]{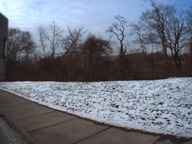} &
        \includegraphics[width=\imwww\textwidth]{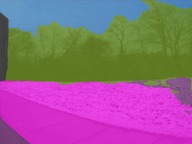} &
        \includegraphics[width=\imwww\textwidth]{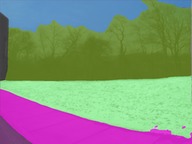} &
        \includegraphics[width=\imwww\textwidth]{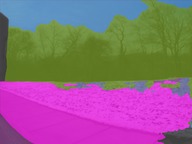} &
        \includegraphics[width=\imwww\textwidth]{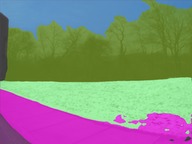} \\
        
        \includegraphics[width=\imwww\textwidth]{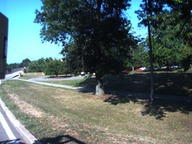} &
        \includegraphics[width=\imwww\textwidth]{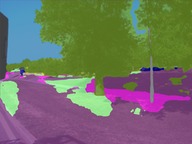} &
        \includegraphics[width=\imwww\textwidth]{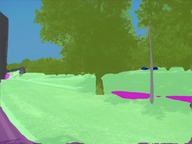} &
        \includegraphics[width=\imwww\textwidth]{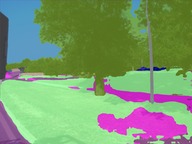} &
        \includegraphics[width=\imwww\textwidth]{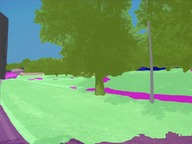} \\
        
    \end{tabular}
    \setlength\tabcolsep{6pt} 
    \caption{Additional qualitative results on unannotated images from the CMU Seasons dataset. Four different networks are compared, the notations used are: E: trained with extra CMU annotations, C: trained with correspondence data, V: trained with Vistas training set.}
    \label{fig:cmu_ims_supp_vis}
\end{figure*}

\def \imwww {0.16}
\newcommand{\supoxima}{dawnleft1418723512039667}
\newcommand{\supoximb}{night-rainrear1418841246916088}
\newcommand{\supoximc}{night-rainright1418841082436378}
\newcommand{\supoximd}{nightrear1418236099050440}
\newcommand{\supoxime}{overcast-summerrear1432294544193098}
\newcommand{\supoximf}{rainrear1416908256629467}
\begin{figure*}
    \centering
    \setlength\tabcolsep{1pt} 
    \begin{tabular}{cccccc}
        Image & Annotation & E & E + C & V + E & V + E + C \\
        \includegraphics[width=\imwww\textwidth]{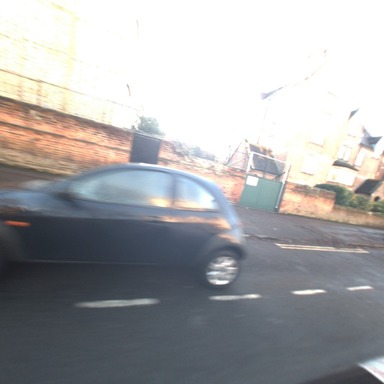} &
        \includegraphics[width=\imwww\textwidth]{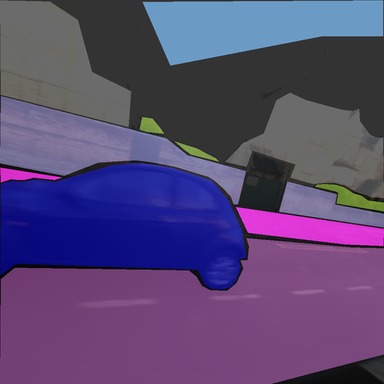} &
        \includegraphics[width=\imwww\textwidth]{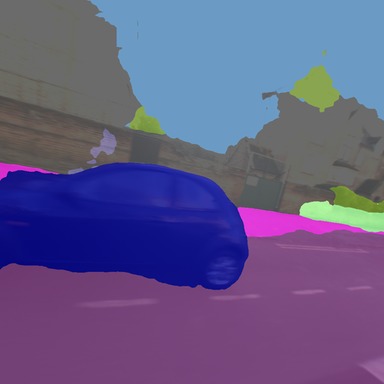} &
        \includegraphics[width=\imwww\textwidth]{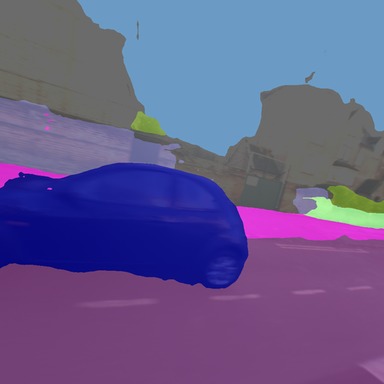} &
        \includegraphics[width=\imwww\textwidth]{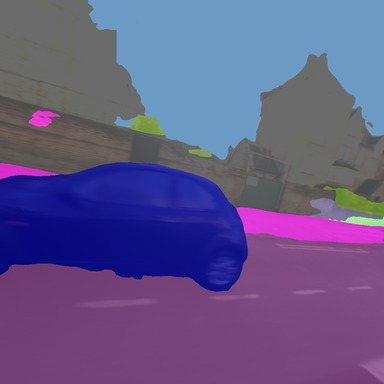} &
        \includegraphics[width=\imwww\textwidth]{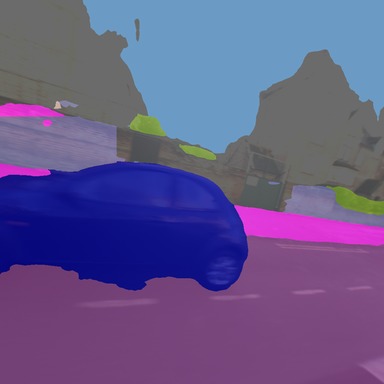} \\
        
        \includegraphics[width=\imwww\textwidth]{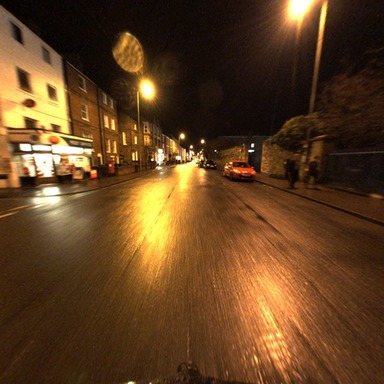} &
        \includegraphics[width=\imwww\textwidth]{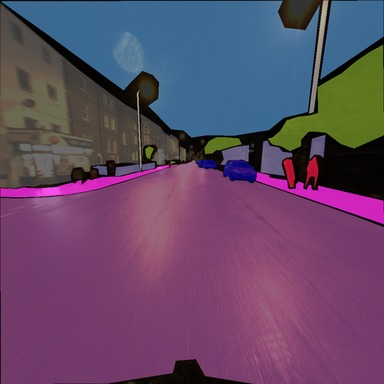} &
        \includegraphics[width=\imwww\textwidth]{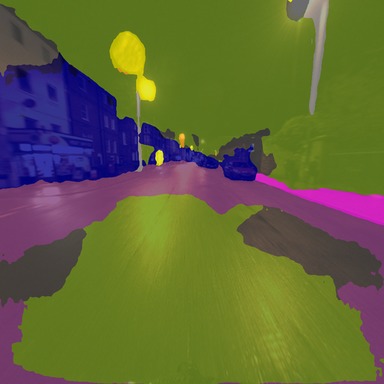} &
        \includegraphics[width=\imwww\textwidth]{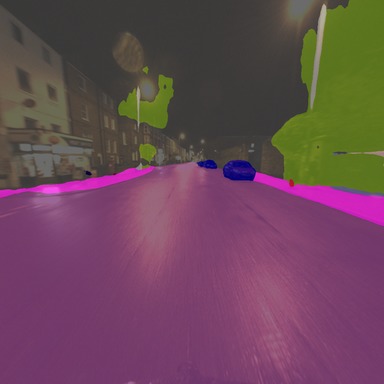} &
        \includegraphics[width=\imwww\textwidth]{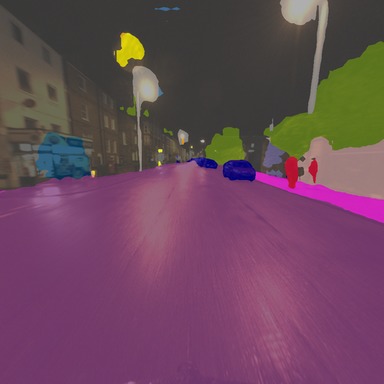} &
        \includegraphics[width=\imwww\textwidth]{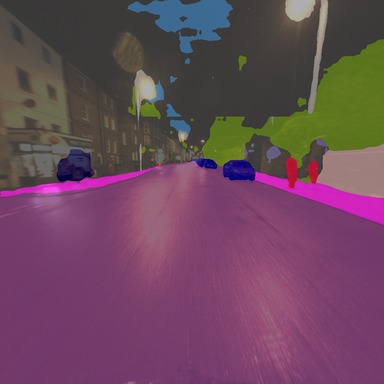} \\
        
        \includegraphics[width=\imwww\textwidth]{figures/results/ox_test_res/\supoximc.jpg} &
        \includegraphics[width=\imwww\textwidth]{figures/results/ox_test_res/\supoximc_truth.jpg} &
        \includegraphics[width=\imwww\textwidth]{figures/results/ox_test_res/\supoximc_cs-extra.jpg} &
        \includegraphics[width=\imwww\textwidth]{figures/results/ox_test_res/\supoximc_cs-extra-corr.jpg} &
        \includegraphics[width=\imwww\textwidth]{figures/results/ox_test_res/\supoximc_cs-map-extra.jpg} &
        \includegraphics[width=\imwww\textwidth]{figures/results/ox_test_res/\supoximc_cs-map-extra-corr.jpg} \\
        
        \includegraphics[width=\imwww\textwidth]{figures/results/ox_test_res/\supoximd.jpg} &
        \includegraphics[width=\imwww\textwidth]{figures/results/ox_test_res/\supoximd_truth.jpg} &
        \includegraphics[width=\imwww\textwidth]{figures/results/ox_test_res/\supoximd_cs-extra.jpg} &
        \includegraphics[width=\imwww\textwidth]{figures/results/ox_test_res/\supoximd_cs-extra-corr.jpg} &
        \includegraphics[width=\imwww\textwidth]{figures/results/ox_test_res/\supoximd_cs-map-extra.jpg} &
        \includegraphics[width=\imwww\textwidth]{figures/results/ox_test_res/\supoximd_cs-map-extra-corr.jpg} \\
        
        \includegraphics[width=\imwww\textwidth]{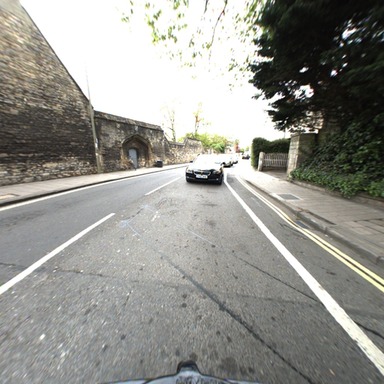} &
        \includegraphics[width=\imwww\textwidth]{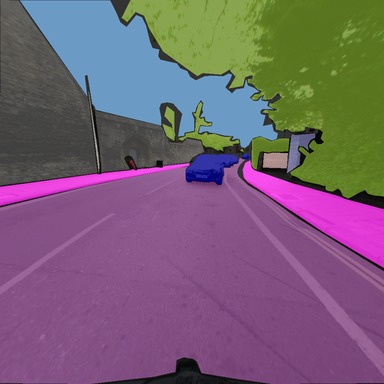} &
        \includegraphics[width=\imwww\textwidth]{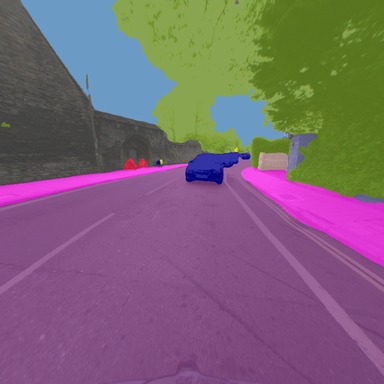} &
        \includegraphics[width=\imwww\textwidth]{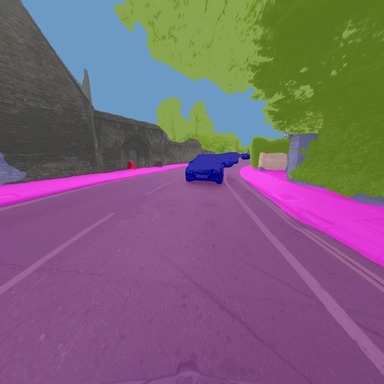} &
        \includegraphics[width=\imwww\textwidth]{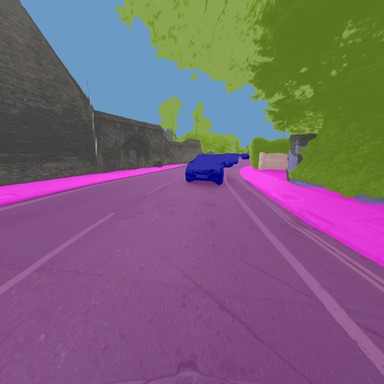} &
        \includegraphics[width=\imwww\textwidth]{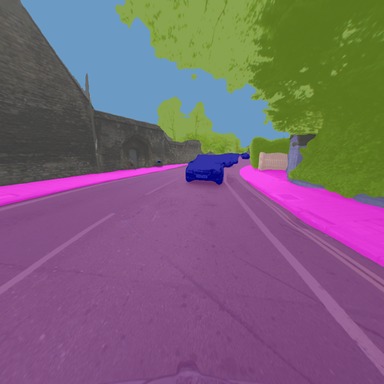} \\
        
        \includegraphics[width=\imwww\textwidth]{figures/results/ox_test_res/\supoximf.jpg} &
        \includegraphics[width=\imwww\textwidth]{figures/results/ox_test_res/\supoximf_truth.jpg} &
        \includegraphics[width=\imwww\textwidth]{figures/results/ox_test_res/\supoximf_cs-extra.jpg} &
        \includegraphics[width=\imwww\textwidth]{figures/results/ox_test_res/\supoximf_cs-extra-corr.jpg} &
        \includegraphics[width=\imwww\textwidth]{figures/results/ox_test_res/\supoximf_cs-map-extra.jpg} &
        \includegraphics[width=\imwww\textwidth]{figures/results/ox_test_res/\supoximf_cs-map-extra-corr.jpg} \\
    \end{tabular}
    \setlength\tabcolsep{6pt} 
    \caption{Qualitative results on the RobotCar Seasons test set. Four different networks are compared, the notations used are: E: trained with extra RobotCar annotations, C: trained with correspondence data, V: trained with Vistas training set.}
    \label{fig:ox_ims_supp}
\end{figure*}

\def \imwww {0.19}

\newcommand{\visoxima}{dawnrear1418721439517734.}
\newcommand{\visoximb}{night-rainleft1418840535628964.}
\newcommand{\visoximc}{night-rainleft1418841246416150.}
\newcommand{\visoximd}{night-rainrear1418841110807877.}
\newcommand{\visoxime}{night-rainrear1418841232167908.}
\newcommand{\visoximf}{nightleft1418237077637177.}
\newcommand{\visoximg}{overcast-summerrear1432293641545821.}
\begin{figure*}
    \centering
    \setlength\tabcolsep{1pt} 
    \begin{tabular}{cccccc}
        Image & E & E + C & V + E & V + E + C \\
        \includegraphics[width=\imwww\textwidth]{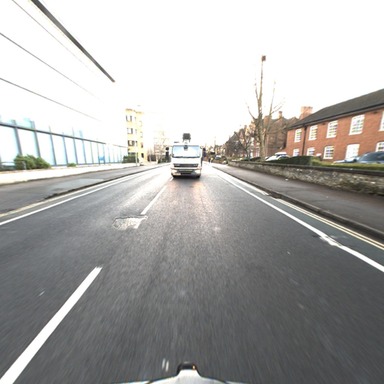} &
        \includegraphics[width=\imwww\textwidth]{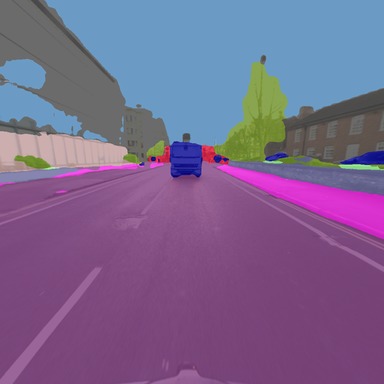} &
        \includegraphics[width=\imwww\textwidth]{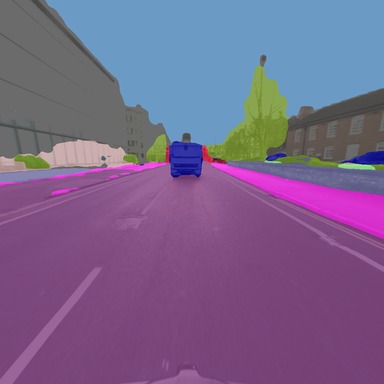} &
        \includegraphics[width=\imwww\textwidth]{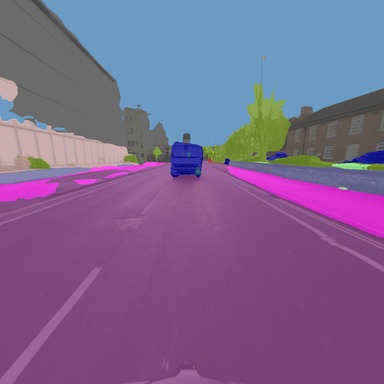} &
        \includegraphics[width=\imwww\textwidth]{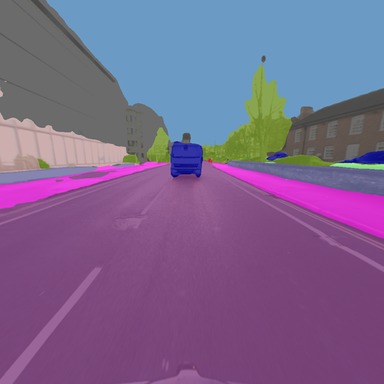} \\

        \includegraphics[width=\imwww\textwidth]{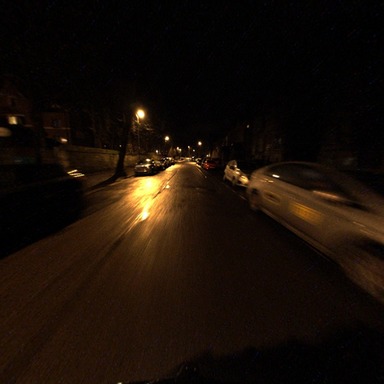} &
        \includegraphics[width=\imwww\textwidth]{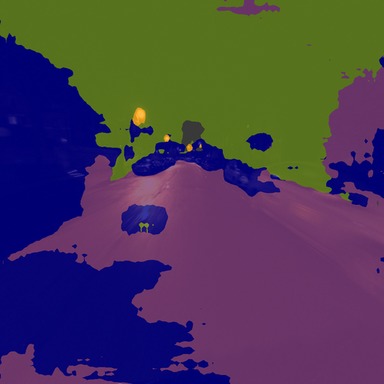} &
        \includegraphics[width=\imwww\textwidth]{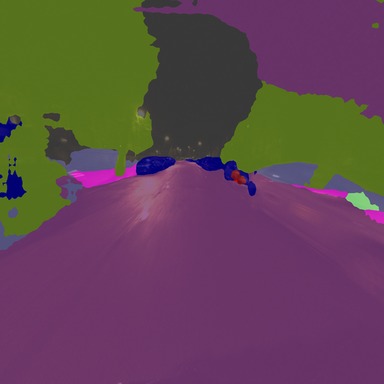} &
        \includegraphics[width=\imwww\textwidth]{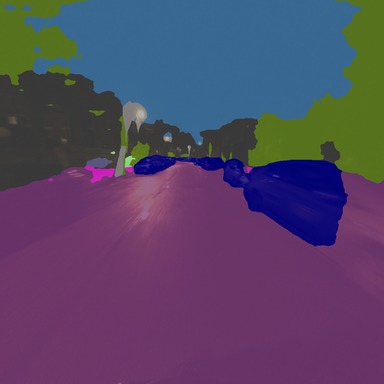} &
        \includegraphics[width=\imwww\textwidth]{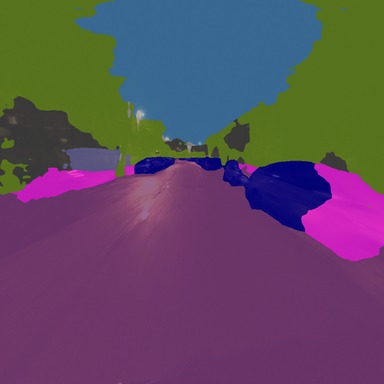} \\
        
        \includegraphics[width=\imwww\textwidth]{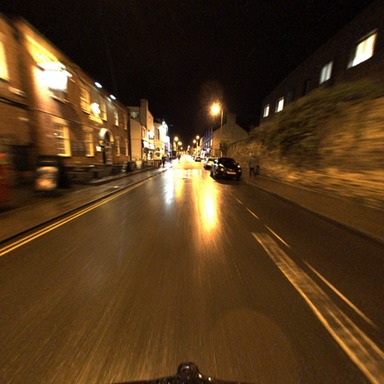} &
        \includegraphics[width=\imwww\textwidth]{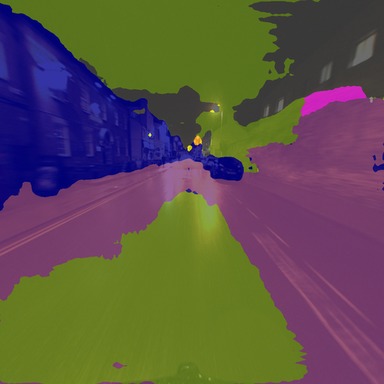} &
        \includegraphics[width=\imwww\textwidth]{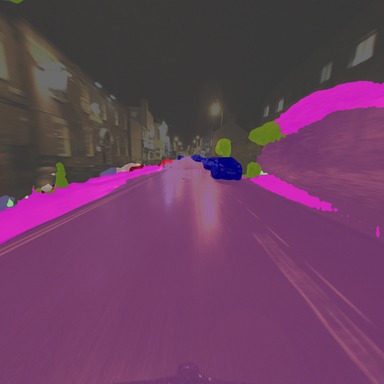} &
        \includegraphics[width=\imwww\textwidth]{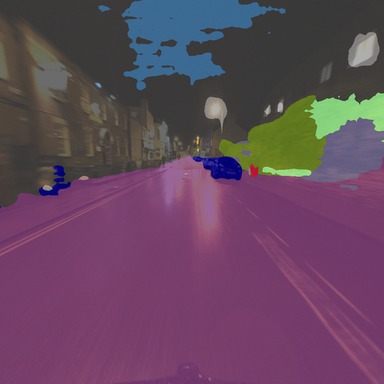} &
        \includegraphics[width=\imwww\textwidth]{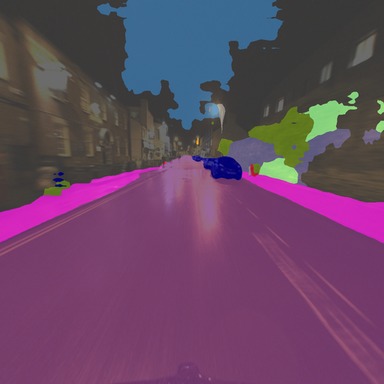} \\
        
        \includegraphics[width=\imwww\textwidth]{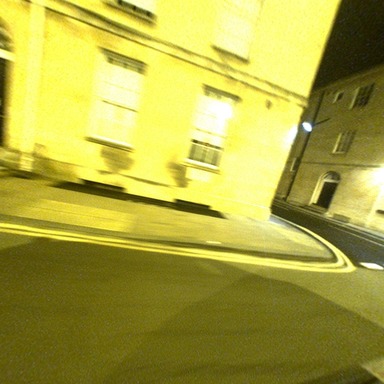} &
        \includegraphics[width=\imwww\textwidth]{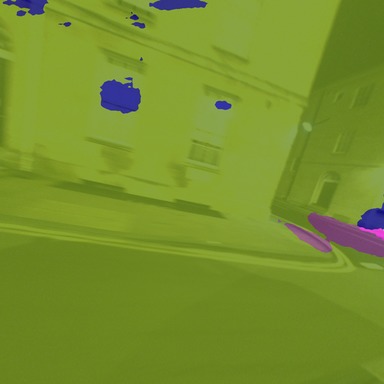} &
        \includegraphics[width=\imwww\textwidth]{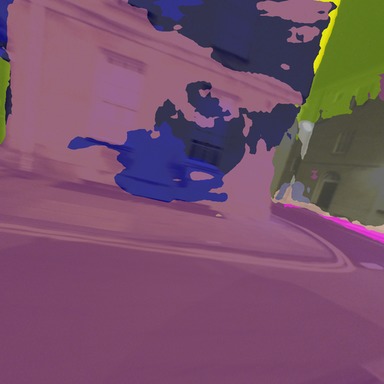} &
        \includegraphics[width=\imwww\textwidth]{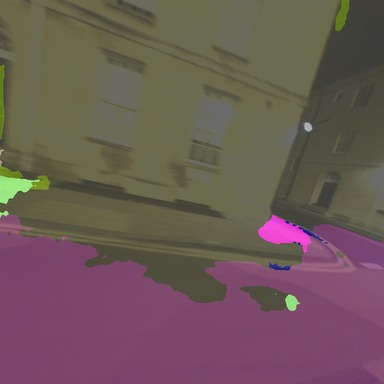} &
        \includegraphics[width=\imwww\textwidth]{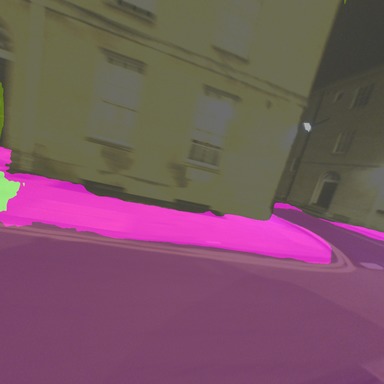} \\
        
        \includegraphics[width=\imwww\textwidth]{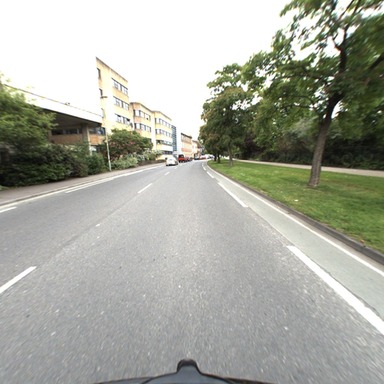} &
        \includegraphics[width=\imwww\textwidth]{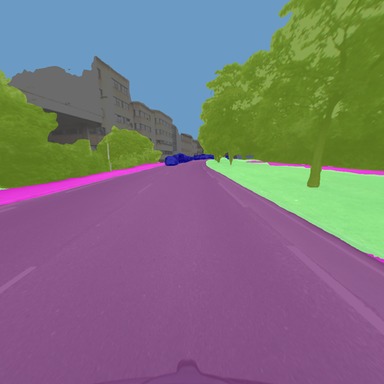} &
        \includegraphics[width=\imwww\textwidth]{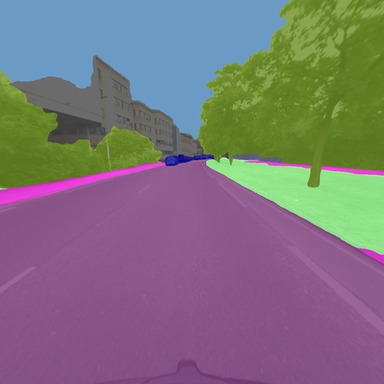} &
        \includegraphics[width=\imwww\textwidth]{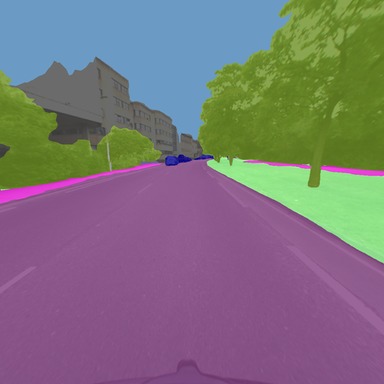} &
        \includegraphics[width=\imwww\textwidth]{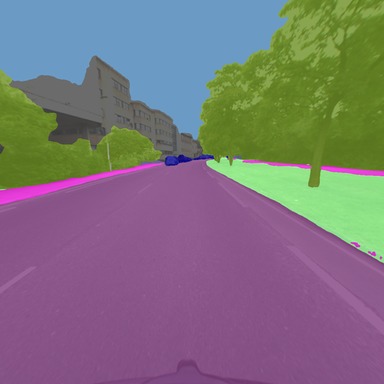} \\
    \end{tabular}
    \setlength\tabcolsep{6pt} 
    \caption{Additional qualitative results on unannotated images from the RobotCar Seasons dataset. Four different networks are compared, the notations used are: E: trained with extra RobotCar annotations, C: trained with correspondence data, V: trained with Vistas training set.}
    \label{fig:ox_ims_supp_vis}
\end{figure*}

\clearpage

{\small
\bibliographystyle{ieee}
\bibliography{mainbib}
}


\end{document}